\PassOptionsToPackage{dvipsnames}{xcolor}
\documentclass[11pt,letterpaper]{article}
\usepackage[margin=1in]{geometry}
\usepackage[T1]{fontenc}
\usepackage{lmodern}
\usepackage[section]{placeins}
\usepackage[utf8]{inputenc}

\usepackage[english]{babel}
\usepackage[round,authoryear]{natbib}
\renewcommand{\bibsection}{\section*{References}}
\usepackage{xcolor}

\usepackage{amsmath,amssymb,amsthm,mathtools}
\usepackage{blkarray,color}
\usepackage{cancel}
\usepackage{enumitem}
\usepackage{euscript}
\usepackage{float}
\usepackage{bm}
\usepackage{graphicx}
\usepackage{array}
\usepackage{tabularx}
\usepackage{booktabs}
\usepackage{mathrsfs}
\usepackage{microtype}
\usepackage{ragged2e}

\usepackage{tikz}
\usepackage[Symbolsmallscale]{upgreek}

\usepackage[font=small,labelfont=bf,skip=6pt]{caption}
\usepackage{hyperref}
\usepackage{url}

\hypersetup{
  colorlinks=true,
  linkcolor=MidnightBlue,
  citecolor=MidnightBlue,
  urlcolor=MidnightBlue,
  pdftitle={Error Accumulation in Second Order Sequential Machine Unlearning},
  pdfauthor={Chen Shang, Haolin Zou, Binghe Zhu, Arian Maleki},
  pdfkeywords={machine unlearning, second-order methods, certified removal, sequential unlearning}
}

\newcommand{\cA}{\mathcal{A}}

\newcommand{\cC}{\mathcal{C}}
\newcommand{\cD}{\mathcal{D}}

\newcommand{\cG}{\mathcal{G}}

\newcommand{\cN}{\mathcal{N}}

\newcommand{\cS}{\mathcal{S}}

\newcommand{\cU}{\mathcal{U}}

\newcommand{\cZ}{\mathcal{Z}}

\newcommand{\EE}{\mathbb{E}}

\newcommand{\II}{\mathbb{I}}

\newcommand{\PP}{\mathbb{P}}

\newcommand{\RR}{\mathbb{R}}

\newcommand{\bs}{\backslash}
\newcommand{\polylog}{\operatorname{polylog}}

\newcommand{\bx}{\bm{x}}
\newcommand{\by}{\bm{y}}

\newcommand*{\triplenorm}[1]{{\left\vert\kern-0.25ex\left\vert\kern-0.25ex\left\vert #1
    \right\vert\kern-0.25ex\right\vert\kern-0.25ex\right\vert}}

\renewcommand{\phi}{\varphi}

\DeclareMathOperator*{\argmin}{argmin}

\renewcommand{\sb}{\beta^*}
\newcommand{\sbk}{\beta_{\bs k}^*}
\newcommand{\tb}{\tilde{\beta}}
\newcommand{\tbk}{\tilde{\beta}_{\bs k}}
\newcommand{\tbkp}{\tilde{\beta}_{\bs k}^{\rm pub}}
\newcommand{\tO}{\tilde{O}}

\newcommand\numberthis{\addtocounter{equation}{1}\tag{\theequation}}

\usepackage[ruled,vlined]{algorithm2e}
\providecommand{\textproc}[1]{\textsc{#1}}

\usepackage[capitalize,noabbrev]{cleveref}

\newtheorem{theorem}{Theorem}[section]
\newtheorem{proposition}[theorem]{Proposition}
\newtheorem{lemma}[theorem]{Lemma}
\newtheorem{corollary}[theorem]{Corollary}
\newtheorem{definition}{Definition}[section]
\newtheorem{assumption}{Assumption}[section]
\newtheorem{remark}{Remark}[theorem]

\title{Error Accumulation in Second-Order Sequential Machine Unlearning}
\author{%
  Chen Shang\textsuperscript{1}\quad Haolin Zou\textsuperscript{1}\quad
  Binghe Zhu\textsuperscript{1}\quad Arian Maleki\textsuperscript{1,2}\\[0.6em]
  {\normalsize\textsuperscript{1}Department of Statistics, Columbia University}\\
  {\normalsize\textsuperscript{2}Department of Electronic Engineering, Columbia University}%
}
\date{}

\makeatletter
\renewcommand{\@maketitle}{%
  \begin{center}
    {\LARGE\bfseries \@title\par}
    \ifx\@author\@empty\else
      \vspace{1.2em}{\large\begin{tabular}[t]{c}\@author\end{tabular}\par}
    \fi
    \ifx\@date\@empty\else\vspace{0.8em}{\@date\par}\fi
  \end{center}
  \vspace{0.8em}
}
\makeatother

\begin{document}
\maketitle
\begin{abstract}

Machine unlearning seeks to remove the influence of deleted training data without full retraining. In the sequential setting, deletion requests arrive over time, and repeated updates can accumulate approximation errors. In this paper, we study error accumulation in high dimensions, where parameter count $p$ is proportional to sample size $n$, for second-order unlearning methods under two privacy settings: (i) \emph{partial unlearning}, which allows unpublished internal states to be retained while certified versions of the model are released; and (ii) \emph{perfect unlearning}, which retains only published models and no hidden states. We also distinguish pointwise and pathwise certification, depending on whether the certification guarantee applies to each released model or the whole released sequence. We analyze conjugate gradient (CG) as an efficient solver for the inverse Hessian-vector product (IHVP) in Newton iterations. Our theory shows that a maximum of 1/2/2/3 Newton--CG updates suffice for successful certification and vanishing population-risk degradation of the published model, for pointwise partial, pathwise partial, pointwise perfect and pathwise perfect certification, respectively.
Numerical experiments demonstrate accurate sequential tracking and substantial computational gains of Newton-based methods combined with conjugate gradient, over gradient-based updates, especially for ill-conditioned problems.
\end{abstract}


\medskip
\section{Introduction}

Machine unlearning aims to remove the influence of deleted training observations without retraining a model from scratch \citep{cao2015towards, ginart2019making}. 
This capability is increasingly important in light of emerging legal requirements, such as the European Union’s General Data Protection Regulation (GDPR) and the California Consumer Privacy Act (CCPA). In many applications, deletion requests arrive sequentially: after each deletion, the updated model becomes the starting point for the next update. Approximate unlearning errors can therefore propagate across deletion requests, making guarantees for a single deletion insufficient for understanding a long unlearning trajectory.

Existing certified unlearning methods provide  distributional indistinguishability to retraining while maintaining a certain accuracy level, often through first or second-order approximations combined with randomization \citep{guo2019certified, neel2021descent, sekhari2021remember}.  Moreover, sequentially updating and releasing the model raises a second problem: certifying each released model separately does not directly certify the joint sequence of releases, which is known as the privacy composition in differential privacy \citep{kairouz2013TheCT}. Moreover, publication noise may itself enter subsequent updates when unpublished model states cannot be kept \citep{neel2021descent}.

We study sequential second-order unlearning in the proportional high-dimensional regime ($p\asymp n$), allowing linear ($T\leq c n$) deletion requests for any fixed proportion $c<1$. The central question is: in order to sequentially delete a certain proportion of the training data, \textbf{how many Newton iterations per deletion are needed to maintain certification and accuracy?} Our contribution is an analysis of sequential error accumulation and its algorithmic consequences under a specified high-dimensional generalized linear model. The analysis connects approximation accuracy, publication-noise feedback, and numerical stopping criteria over a linear deletion horizon.
\begin{itemize}
    \item \textbf{Partial unlearning: uniform error tracking}. We analyze how deletion shifts, inherited error from the previous deletion, and an inexact CG solution affect error accumulation over a linear deletion horizon. For partial unlearning initiated at a hidden model state, one Newton step per deletion suffices for pointwise certification and vanishing population-risk degradation. An additional Newton step achieves pathwise certification.
    \item \textbf{Perfect unlearning: publication-noise feedback.} 
    In perfect unlearning where only the published model may be retained and used for the next deletion, publication noise is also fed back into the error of subsequent deletions. We control its effect on the Newton updates even when the Euclidean norm of noises does not vanish. With the noise levels and error tolerances specified, two Newton steps suffice in the pointwise setting and three in the pathwise setting, with high probability. Our sharpness analysis also shows that the 1/2/2/3 Newton steps for the four cases cannot be reduced for the prescribed Newton update rule and parameter choices.
    \item \textbf{Sequential guarantees for Newton--CG updates.} To avoid explicit Hessian inversion, we use matrix-free conjugate gradient (CG) for efficient inverse-hessian-vector-product (IHVP) computation, and theoretically quantify how much error from the extra CG approximation can preserve the sequential guarantees using exact hessian inverses. Experiments on high-dimensional logistic regression show that Newton-CG can substantially reduce computation relative to gradient-based updates, even for problems with ill-conditioned covariate covariances. 
\end{itemize}

Our work builds most directly on certified removal via Newton approximations \citep{guo2019certified}, sequential certified unlearning \citep{neel2021descent,gupta2021adaptive, suriyakumar2022algorithms}, and deletion-capacity analyses \citep{sekhari2021remember}. Notably, \cite{neel2021descent} distinguished settings according to whether unpublished internal state can be maintained using gradient-based methods,
but their general bounds do not directly establish the accuracy guarantees sought here in the high-dimensional regime.  \citet{zou2025certified,pandey2025gaussian} study Newton-based certified unlearning in the proportional high-dimensional regime. Both works consider a single removal batch of deletions, starting from the original fitted model, whereas our sequential problem starts each deletion from the previous approximate center or even noisy published model.  In our work, we study the sequential problem and control error accumulation uniformly over a linear number of deletions. Our computational results are also related to recent efforts to make second-order unlearning scalable, such as \cite{qiao2025hessian}. Work on neural-network and LLM unlearning \citep{koloskova2025certified, bui2026how} and on unlearning-friendly training architectures \citep{bourtoule2021machine} is complementary to our focus on rigorous sequential guarantees for regularized ERM.
\section{Framework and Preliminaries}

\subsection{Sequential Unlearning Framework}

Suppose we observe a dataset $\cD=\{z_i\}_{i=1}^n$, with each observation $z_i\in\cZ$ for some data domain $\cZ$. For simplicity, we assume the datapoints $z_1, z_2, \dots$ are deleted in ascending order, and write $\cD_{\backslash k}=(z_{k+1},\ldots,z_n)$
for the dataset retained after the first $k$ deletions,
with $\cD_{\backslash 0}=\cD$.

We adopt the unlearning framework in \cite{neel2021descent} with a slight modification. A \textit{learning algorithm} is a mapping $\cA: \left(\bigcup_{m=1}^n\cZ^m\right)\to\Theta$ that maps the dataset $\cD\in\cZ^n$ or its subsets to a learned model $\sb\in\Theta$. In this paper, we consider the regularized empirical risk minimization (R-ERM) as the learning algorithm:
\[
    \begin{aligned}
    \sb:=\mathcal{A}(\mathcal{D})&:=\argmin_{\beta\in\mathbb R^p} \sum_{i=1}^{n}\ell(y_i, x_i^\top\beta)+\lambda r(\beta),\\
    \sbk:=\mathcal{A}(\mathcal{D}_{\backslash k})&:=\argmin_{\beta\in\mathbb R^p} \sum_{i=k+1}^{n}\ell(y_i, x_i^\top\beta)+\lambda r(\beta).
    \end{aligned}\numberthis\label{eq: exact retraining}
\]
Note that the regularization strength $\lambda>0$ is fixed for all $n$ and all deletion steps $k\leq T$.

An \textit{unlearning algorithm} is a mapping $\cU: \left(\bigcup_{m=1}^n\cZ^m\right)\times \cZ \times \Theta\longrightarrow \Theta$ which takes three components as inputs: the dataset $\cD\in \cZ^n$ or its subset before deletion, a data point $z_k\in\cD$ to be deleted, and the model $\beta\in\Theta$ before deletion, and outputs an updated model $\tb$ that is supposed to remove all influence from the deleted data point $z_k$.  To protect private information of the deleted point $z_k$, the output of the learning algorithm $\cA(\cD)$ and unlearning algorithm $\cU(\cD, z, \beta)$ should go through a \textit{publishing function} $f_{\rm pub}:\Theta \to\Theta$ before being released to the public. Throughout this paper, the output of the publishing function $f_{\rm pub}$ is considered as a \textbf{randomized} version of its input, e.g. perturbing the input model with a suitable noise. 

One objective of machine unlearning is called \textit{certification} (\cite{guo2019certified, neel2021descent}) which guarantees statistical indistinguishability between the published versions of the unlearned model and the exactly retrained model.
\begin{definition}[($\epsilon, \delta$)-indistinguishability]\label{def: eps delta indistinguishability}
Two random vectors $\bx, \by$ defined on some domain $\cS$ are said to be ($\epsilon, \delta$)-indistinguishable, written as $\bx \overset{\epsilon, \delta}{\approx}\by$, iff $\forall$ measurable set $S\subset \cS$,  
\begin{align*}
    \mathbb{P}(\bx\in S) &\leq {\rm e}^{\epsilon}\mathbb{P}( \by \in S) + \delta,\\
    \mathbb{P}(\by\in S) &\leq {\rm e}^{\epsilon}\mathbb{P}(\bx\in S ) + \delta.
\end{align*}

\end{definition}

\begin{definition}[($\epsilon, \delta$)-pointwise certification]\label{def: pointwise unlearning}
Let $\cA$ and $\cU$ be two learning and unlearning algorithms. Let $\tilde{\beta}_{\backslash k}=\cU(\cD_{\backslash k-1}, z_k, \tilde{\beta}_{\backslash k-1})$ be defined recursively with $\tilde{\beta}_{\backslash 0} = \cA(\cD)$. We say the pair $(\cU, f_{\rm pub})$ satisfies \textbf{($\epsilon, \delta$)-pointwise partial certification} iff  $\forall \cD$, $\forall 1\leq k\leq T$, 
\[
    f_{\rm pub}(\cU(\cD_{\backslash k-1}, z_k, \tilde{\beta}_{\backslash k-1})) \overset{\epsilon, \delta}{\approx} f_{\rm pub} (\cA(\cD_{\backslash k})).
\]
Moreover, if we define $$\tilde{\beta}^{\rm pub}_{\backslash k}:=f_{\rm pub} (\cU(\cD_{\backslash k-1}, z_k, \tilde{\beta}^{\rm pub}_{\backslash k-1}))$$ recursively with $\tilde{\beta}^{\rm pub}_{\backslash 0} = \cA(\cD)$, then $(\cU, f_{\rm pub})$ is said to satisfy \textbf{($\epsilon, \delta$)-pointwise perfect certification} iff $\forall \cD$, $\forall 1\leq k\leq T$
\[
     \tilde{\beta}^{\rm pub}_{\backslash k}\overset{\epsilon, \delta}{\approx} f_{\rm pub} (\cA(\cD_{\backslash k})).
\]
\end{definition}

\begin{definition}[($\epsilon, \delta$)-pathwise certified unlearning]\label{def: pathwise unlearning}
    Using the same notations as Definition \ref{def: pointwise unlearning}, we say $(\cU, f_{\rm pub})$ satisfies \textbf{($\epsilon, \delta$)-pathwise partial certification} iff  $\forall \cD$, 
    \begin{multline*}
        \bigl(f_{\rm pub}(\cU(\cD, z_1, \tilde{\beta}_{\backslash 0})),\dots,
        f_{\rm pub}(\cU(\cD_{\backslash (T-1)}, z_T, \tilde{\beta}_{\backslash T-1}))\bigr)\\
        \overset{\epsilon, \delta}{\approx} (f_{\rm pub} (\cA(\cD_{\backslash 1})),\dots, f_{\rm pub} (\cA(\cD_{\backslash T}))) .
    \end{multline*}
    Moreover, $(\cU, f_{\rm pub})$ is said to satisfy \textbf{($\epsilon, \delta$)-pathwise perfect certification} iff  $\forall \cD$,  
    \[
        \begin{gathered}
        (\tilde{\beta}^{\rm pub}_{\backslash 1}, \dots, \tilde{\beta}^{\rm pub}_{\backslash T})
        \overset{\epsilon, \delta}{\approx} (f_{\rm pub} (\cA(\cD_{\backslash 1})),\dots, f_{\rm pub} (\cA(\cD_{\backslash T}))) .
        \end{gathered}
    \]

\end{definition}
Intuitively, a \textit{partially certified} unlearning algorithm starts its iteration from the unpublished (thus uncertified) model state of the last deletion, $\tilde{\beta}_{\backslash k}$. The perfect certification does not allow the owner of the model to keep the unpublished sequence $\tilde{\beta}_{\backslash k}$, so they must directly use the published models $\tilde{\beta}^{\rm pub}_{\backslash k-1} = f_{\rm pub}(\tilde{\beta}_{\backslash k-1})$. Since the published version usually contains more noise than the unpublished version, perfect unlearning is more stringent and may require additional computation resources.

Pointwise certification controls the privacy leakage of each released model separately, but does not by itself control the whole release sequence. Pathwise certification mitigates the issue by guaranteeing  indistinguishability over the whole published model sequence.
\paragraph{Certification on a data class.} Theoretical guarantees may not cover all possible $\cD\in\cZ^n$ due to the existence of extreme or pathological realizations of $\cD$, but with reasonable assumptions the theory should apply to a large enough data class $\mathcal G_n \subseteq \mathcal Z^n$ such that $\cD\in\mathcal G_n$ with high probability. Consequently, we say that $(U,f_{\mathrm{pub}})$ satisfies any of the
certification notions in Definitions \ref{def: pointwise unlearning}-\ref{def: pathwise unlearning}
\emph{on $\mathcal G_n$} if the corresponding distributional
comparisons hold for every fixed $\cD \in \mathcal G_n$,
with the same parameters $(\epsilon,\delta)$.
Taking $\mathcal G_n = \mathcal Z^n$ recovers the unrestricted definitions. Appendix~\ref{app:data-event} constructs a
data-only measurable class $\mathcal G_n$ under mild assumptions on the data distribution, such that $\mathcal G_n$ almost covers $\mathcal Z_n$ with near zero failing probability:
\begin{equation}\label{eq:good-data-probability-main}
    \mathbb P_D(D\notin\mathcal G_n)
    \le p_n
    := n q_y(n)+n^{-d}
    = o(1).
\end{equation}
The certification results in
Section~\ref{sec:main-theory} hold on $\mathcal G_n$ in the data-class
sense specified after Definition~\ref{def: pathwise unlearning}.


Essentially any unlearning method can be certified by a large enough perturbation in the publication function $f_{\rm pub}$, but such a large noise might harm the performance of the model in subsequent tasks. Therefore we need an accuracy measure for the published model $\tbkp$.
\begin{definition}[Population-risk degradation]\label{def:prd}\label{def:ged}
Let $R(\beta)=\mathbb E_{(x,y)}[\ell(y,x^\top\beta)]$, where $(x,y)$
is an independent observation from the training population. Define
\[
 \begin{aligned}
 \operatorname{PRD}_k
 &:=\bigl[R(\widetilde\beta^{\rm pub}_{\backslash k})
                  -R(\beta_{\backslash k}^*)\bigr]_+,
 \\
 \operatorname{PRD}_{1:T}&:=\max_{1\leq k\leq T}\operatorname{PRD}_k,
 \end{aligned}
\]
where $[a]_+=\max\{a,0\}$.
\end{definition}

 Various accuracy measures have been used in the literature, including the excess risk in \cite{sekhari2021remember} that compares the population risk between $\tbkp$ and a "true parameter".  Our population-risk degradation (PRD) instead compares the population risk against the exactly retrained model $\sbk$, without including the statistical error against population optimum already incurred by training or retraining.

\subsection{Newton--CG unlearning}\label{sec:newton-cg-main}
Suppose the objective $F_{\bs k}$ is $\mu$-strongly convex.
At each deletion $k$, the goal is to approximately find $\sbk=\cA(\cD_{\bs k})=\argmin F_{\bs k}$ and certify it through a proper publication function.

Define $g_{\bs k}=\nabla F_{\bs k}\in\RR^p$ and $H_{\bs k}=\nabla^2F_{\bs k}\in\RR^{p\times p}$.
The exact Newton mapping is defined as
\begin{equation}\label{eq: sequential unlearning estimate beta tilde k}
 \mathcal N_k(u)=u-H_{\bs k}(u)^{-1}g_{\bs k}(u).
\end{equation}
The Newton-Raphson method recursively applies \eqref{eq: sequential unlearning estimate beta tilde k}, until a solution $\tbk\approx \sbk$ is found. Compared to first-order methods such as stochastic gradient descent (SGD), Newton's method uses the inverse Hessian instead of a scalar step size, so it saves the effort of tuning it.  It is known to have rapid convergence under mild conditions, at the sacrifice of computation cost. In this section we will present our Newton unlearning algorithms, under partial or perfect unlearning, that satisfy pointwise or pathwise certification. We will also use the conjugate gradient (CG) descent to address the computation issue.


\paragraph{Newton partial unlearning}
We start with partial unlearning, where the system is allowed to retain a hidden model state for the next deletion. 

For the initialization, for partial unlearning we can retain the unpublished model from the last deletion, $\tb_{\bs (k-1)}$, as initialization for the current deletion. For the stopping rule, suppose we want to solve $\sbk$ up to some precision $\tau_n>0$, that is, finding $\tbk$ such that $\|u-\beta^*_{\backslash k}\|_2\leq\tau_n$. By strong convexity, 
\begin{equation}\label{eq:perfect-gradient-test}
 \|g_{\backslash k}(u)\|_2\leq\mu\tau_n\Rightarrow \|u-\beta^*_{\backslash k}\|_2\leq\tau_n,
\end{equation}
where $\mu=\lambda\nu$ is the strong convexity constant of $F_{\bs k}$. So it is sufficient to control the \textit{gradient residual} $\|g_{\backslash k}(u)\|_2\leq\mu \tau_n$. In Section \ref{ssec: partial unlearning theory} we will show that, in most cases,  one or two Newton iterations \eqref{eq: sequential unlearning estimate beta tilde k} suffice for pointwise and pathwise certification respectively.

\paragraph{Certification via Gaussian noise}
After a satisfactory $\tbk$ is found, we will publish the model by adding a Gaussian noise
\[
    \tbkp=f_{\rm pub}(\tbk) = \tbk + N(0,\sigma_n^2\II_p),
\]
independently drawn for each deletion. For a fixed dataset and deletion order, suppose that
\(\|\widetilde{\beta}_{\backslash k}-\beta^\star_{\backslash k}\|_2\le\tau_n\)
for every \(k\le T\). By Theorem \ref{thm:gaussian-certification}, for \(0<\epsilon\le1\) and \(0<\delta<1/2\), it suffices to set $\sigma_n \gtrsim \frac{\tau_n\sqrt{\log(1/\delta)}}{\epsilon}$
for pointwise certification, and $\sigma_n \gtrsim\frac{\tau_n\sqrt{T\log(1/\delta)}}{\epsilon}$ for pathwise certification.
Details of the noise level will be discussed in Section \ref{sec:main-theory}.


\paragraph{Newton perfect unlearning}
Unlike partial unlearning, the perfect unlearning does not allow us to retain the unpublished models $\tb_{\bs (k-1)}$. Instead, we initialize the Newton iteration at the previous published model $\tb_{\bs (k-1)}^{\rm pub}$. Intuitively, since $\tb_{\bs (k-1)}^{\rm pub}$ is a noisy version of $\tb_{\bs (k-1)}$, it is in general a worse initialization and more Newton steps are required. When publication noise enters the initialization of the next deletion, we jointly choose the center tolerance and the noise scale to control sequential error propagation. In particular, for pointwise perfect unlearning, we use a tighter center tolerance to permit a smaller publication-noise variance. Under the noise levels and center tolerances specified in Section \ref{ssec: perfect unlearning theory}, two Newton-CG steps suffice in the pointwise setting and three in the pathwise setting, with high probability along the deletion sequence.

\paragraph{Conjugate gradient method}
The Newton update computes $H_{\bs k}(u)^{-1}g_{\bs k}(u)$, which is also called an inverse-hessian-vector-product (IHVP) operation. A direct dense factorization to compute the IHVP in general costs $O(p^3)$, which is prohibitive for large models. The conjugate gradient (CG) method \citep{hestenes1952methods} is a popular choice for estimating the IHVP efficiently. We use a zero-start conjugate gradient (\textproc{ConjugateGrad}, Algorithm~\ref{alg:cg}) to solve the linear system $H_{\bs k}(u)d=g_{\bs k}(u)$ up to a small tolerance $\eta>0$ such that $\|g_{\bs k}(u)-H_{\bs k}(u)d\|_2\leq \eta$  (\textproc{ConjugateGrad}, Algorithm~\ref{alg:cg}) for each Newton step. The choice of $\eta$ balances both computation cost (number of CG iterations) and accuracy in solving the Newton direction. Detailed discussion is postponed to Section \ref{sec:main-theory}. In subsequent discussions, define $\cN_k^{\rm CG}(u)$ to be the Newton--CG update, where $d$ is an approximate solution for $H_{\bs k}(u)^{-1}g_{\bs k}(u)$ using \textproc{ConjugateGrad}. Specifically, Algorithm \ref{alg:cg} returns the first iterate $d$ satisfying 
\[
    \|g_{\backslash k}(u)-H_{\backslash k}(u)d\|_2 \leq \frac{\mu\tau_n}{8}
\]
If $g_{\backslash k}(u)=0$, we set $d=0$ without running CG.
Define $\mathcal N_k^{\mathrm{CG}}(u)=u-d$.
Strong convexity gives
\[
  \|\mathcal N_k^{\mathrm{CG}}(u)-\mathcal N_k(u)\|_2
  \leq \frac{\tau_n}{8}.
\]
Let $J_{*}$ denote the prescribed sufficient number
of Newton--CG updates for the selected unlearning setting.
All four settings use this residual rule with their prescribed
center tolerance $\tau_n$.
Section~\ref{sec:main-theory} specifies the center tolerances
and publication-noise scales;
Appendix~\ref{app:cg-proof} analyzes the residual rule
and CG arithmetic work.



\paragraph{Fallback procedure}

We check the gradient gate after every Newton--CG update and stop as soon as it is satisfied. If it still fails after $J_*$ updates, we apply a fallback procedure dubbed $\operatorname{Recover}_k$ to continue optimization on the
retained data until $\|g_{\backslash k}(v)\|_2\leq\mu\tau_n$.
$\operatorname{Recover}_k$ uses gradient descent with Armijo backtracking to select step size from objective evaluations without a known global smoothness
constant. Appendix~\ref{app:recovery-procedure} specifies the procedure
in detail and proves finite termination at this positive tolerance under our
assumptions in exact arithmetic. All failed Newton attempts and
recovery work are included in the computational cost.

Algorithm \ref{alg:sequential-newton-cg} summarizes the unlearning procedure described in this section.

\begin{algorithm}[t]
\small
\DontPrintSemicolon
\caption{Precision-stopped sequential unlearning with recovery}
\label{alg:sequential-newton-cg}
\KwIn{Dataset $D$; exact full-data fit $\beta^\star$; deletion sequence
  $z_{1:T}$; maximum Newton steps $J_*\in\{1,2,3\}$;
  precision $\tau_n>0$;
  noise level $\sigma_n>0$;  Boolean flag \texttt{is\_perfect}.}

\eIf{\texttt{is\_perfect} is \texttt{false}}{
  Set $\widetilde\beta_{\setminus0}\leftarrow\beta^\star$\;
}{
  Set $\widetilde\beta^{\mathrm{pub}}_{\setminus0}
    \leftarrow\beta^\star$\;
}

\For{$k=1,\ldots,T$}{
  Remove $z_k$ and form $F_{\setminus k}$ on $D_{\setminus k}$\;

  \eIf{\texttt{is\_perfect} is \texttt{false}}{
    Set $u\leftarrow\widetilde\beta_{\setminus(k-1)}$\;
  }{
    Set $u\leftarrow\widetilde\beta^{\mathrm{pub}}_{\setminus(k-1)}$\;
  }

  Set $j\leftarrow1$ and \texttt{passed}$\leftarrow$\texttt{false}\;

  \While{\texttt{passed} is \texttt{false}}{
    \eIf{$j\leq J_*$}{
      $u\leftarrow\mathcal N_k^{\mathrm{CG}}(u)$;
      $j\leftarrow j+1$\;
    }{
      $u\leftarrow\operatorname{Recover}_k(u)$\;
    }
    \texttt{passed}$\leftarrow
      [\|g_{\setminus k}(u)\|_2\le\mu\tau_n]$\;
  }

  Draw fresh $Z_k\sim\mathcal N(0,\sigma_n^2I_p)$, independently\;

  Set $\widetilde\beta^{\mathrm{pub}}_{\setminus k}\leftarrow u+Z_k$
    and publish $\widetilde\beta^{\mathrm{pub}}_{\setminus k}$\;

  \eIf{\texttt{is\_perfect} is \texttt{false}}{
    Set $\widetilde\beta_{\setminus k}\leftarrow u$ and retain it as the
      model state for the next deletion\;
  }{
    Retain $\widetilde\beta^{\mathrm{pub}}_{\setminus k}$ as the sole carried
      model state; keep the retained dataset $D_{\setminus k}$ and discard
      $u$, $Z_k$, and all other model and solver states\;
  }
}
\end{algorithm}

%

\section{Main theoretical contributions}
\label{sec:main-theory}
\begin{table}[!htbp]
\centering
\caption{Certification and accuracy guarantees in Theorems~\ref{thm:hidden-one-step}--\ref{thm:public-three-step},
for fixed $(\epsilon,\delta)$. On the ``good data class'' $\cG_n$, each deletion
achieves the desired accuracy $\tau_n$ within $J_\star$ Newton steps.}
\label{tab:sequential-guarantees}
\begingroup
\footnotesize
\setlength{\tabcolsep}{4pt}
\renewcommand{\arraystretch}{1.2}
\begin{tabular*}{\textwidth}{@{\extracolsep{\fill}}l l c c c c@{}}
\toprule
Hidden state & Certification & \shortstack{Center Tolerance \\ $\tau_n$} & \shortstack{Max Newton-CG \\steps $J_*$}
& \shortstack{Noise std\\per coordinate}
& \shortstack{Max risk along\\ deletion path $\operatorname{PRD}_{1:T}$} \\
\midrule
\shortstack[l]{Allowed\\(partial unlearning)} & Pointwise &$\widetilde{O}(n^{-1/2})$ & $1$
& $\widetilde O(n^{-1/2})$ & $\widetilde O_p(n^{-1})$ \\
& Pathwise & $\widetilde{O}(n^{-1})$ & $2$
& $\widetilde O(\sqrt{T}/n)$ & $\widetilde O_p(\sqrt{T}/n^{3/2})$ \\
\midrule
\shortstack[l]{Forbidden\\(perfect unlearning)} & Pointwise & $\widetilde{O}(n^{-1})$& $2$
& $\widetilde O(n^{-1})$ & $\widetilde O_p(n^{-3/2})$ \\
& Pathwise & $\frac{1}{n}$ & $3$
& $\tO(\sqrt{T}/n)$ & $\widetilde O_p(\sqrt{T}/n^{3/2})$ \\
\bottomrule
\end{tabular*}

\smallskip
\begin{minipage}{\textwidth}
\footnotesize

\end{minipage}
\endgroup
\end{table}
Before delving into the assumptions and theorems, Table \ref{tab:sequential-guarantees} summarizes our main theoretical results. The four rows correspond to the four unlearning requirements in Definitions  \ref{def: pointwise unlearning} and \ref{def: pathwise unlearning}. With high probability, our theory shows that the four cases require only 1/2/2/3 Newton--CG iterations to achieve a proper accuracy level (before publication) $\|\tbk-\sbk\|\leq \tau_n$, for some theory-guided tolerance $\tau_n$. Independent Gaussian noises are added to $\tbk$ to guarantee certification, and the noise level $\sigma_n$ is chosen according to $\tau_n$. The required Newton--CG steps $J_*$ guarantee $\tau_n$ and $\sigma_n$ are small enough in each case, so that the published (noisy) models $\tbkp$ still have vanishing risk, measured by population-risk degradation (PRD, Definition \ref{def:prd}).


Table~\ref{tab:sequential-guarantees} summarizes the resulting tradeoffs
among maximum Newton steps, publication noise, and
population-risk degradation under the respective calibrations of
Theorems~\ref{thm:hidden-one-step}--\ref{thm:public-three-step}.

\subsection{Assumptions}\label{sec:assumptions-main}
We use the summed objective in \eqref{eq: exact retraining}, with fixed
$\lambda>0$, $p/n\to\gamma\in(0,1)$, and any prespecified deletion horizon
$1\leq T\leq K:=\lfloor c'n\rfloor$, where $0<c'<1$ is fixed.
The retained sample size is at least $(1-c')n$.
We postpone the exact definitions of our assumptions to Appendix \ref{sec: assumptions} (Assumptions~\ref{ass:design}--\ref{ass:Penalty}). Briefly speaking, the assumptions are:
\begin{itemize}[leftmargin=*,itemsep=1pt]
\item \textbf{A1 (Design).}
Observation pairs $(y_i,x_i)$ are independent draws from a common population,
where $x_i\sim N(0,\Sigma/p)$ and
$\Sigma$ has $\Theta(1)$ eigenvalues.
\item \textbf{A2 (Responses).}
The response variables $y_i$ are stochastically bounded, up to $\polylog(n)$ factors.

\item \textbf{A3 (Loss).}
The loss $\ell(y,z)$ is non-negative,  convex and $\cC^3$. Its derivatives exhibit polynomial growth in both $y$ and $z$.

\item \textbf{A4 (Penalty).}
The penalty $r(\beta)$ is $\cC^2$, separable and
$\nu$-strongly convex with $r(0)=0$ and $\nabla r(0)=0$. Its hessian $\nabla^2r$ is $\polylog(n)$-Lipschitz in Frobenius norm and $\|\nabla^2r\|_{\rm op}=\polylog(n)$.

\end{itemize}
We write
$\mu=\lambda\nu$ to be the strong convexity constant for $F_{\bs k},\forall k\leq T$, and $\widetilde O$ suppresses deterministic fixed powers
of $\log(n)$. 

The assumptions above are mild and include many classic models such as ridge and logistic regression. Assumptions A1 and A2 are conventional in high dimensional statistics. Suppose the population level "true parameter" $\beta^{*,{\rm true}}$ has $O(1)$ coordinates on average, then Assumptions A1 and A2 essentially guarantee $\|x_i\|_2, x_i^\top\beta^{\rm true}$ and $y_i$ are $\tO_p(1)$. Assumptions A3 and A4 are regularity conditions on the loss $\ell$ and penalty $r$ that are conventional in the literature, such as smoothness, strong convexity and Hessian-lipschitzness. The additional polynomial bound on $\ell$ and its derivatives prevents steep variations when the parameter changes, which is important in deciding the perturbation level.

Throughout this section, we use the center tolerances $\tau_n^{(1)}$ and $\tau_{n}^{(2)}$ specified in Appendix \ref{app:hidden-proof}, \cref{eq:two-step-center-tolerance}. They satisfy
\begin{align*}
    \tau_{n}^{(1)} = \widetilde{O}(n^{-1/2}),\quad \tau_{n}^{(2)} = \widetilde{O}(n^{-1}).
\end{align*}

\subsection{Partial unlearning}\label{ssec: partial unlearning theory}
We start with partial unlearning with pointwise certification.
\begin{theorem}[Pointwise partial certification]
\label{thm:hidden-one-step}
Suppose Assumptions~\ref{ass:design}--\ref{ass:Penalty} hold. Use Algorithm~\ref{alg:sequential-newton-cg} with 
${\rm is\_perfect}= {\rm false}, J_\star=1$, $\tau_n=\tau_n^{(1)}$, and
\[
 \sigma_n=2\tau_n^{(1)}\sqrt{2\log(1/\delta)}/\epsilon
\]
with $0<\epsilon\le 1, \delta\in(0,\frac12]$ fixed. For all sufficiently large $n$, the following hold for all $D\in\cG_n$ which has probability at least $1-o(1)$:
\begin{enumerate}
    \item $\underset{k\leq T}{\max}\|\widetilde{\beta}_{\bs k}-\beta^*_{\bs k}\|_2\leq\tau_n^{(1)}$ within $J_*=1$ Newton--CG step,
    \item The published sequence $\{\tbkp\}_{k=1}^T$ satisfies
    $(\epsilon,\delta)$-pointwise partial certification.
    \item For an arbitrary constant $d>1$,$\PP(\operatorname{PRD}_{1:T}>n^{-1}\polylog(n)|\cD)\leq n^{-d}$.
\end{enumerate}
\end{theorem}
\begin{remark}
    In fact, our proof for Theorem \ref{thm:hidden-one-step} shows that, in the case of pointwise partial unlearning, for all sufficiently large $n$ and every $D\in\mathcal{G}_n$, $J_*=1$ \textbf{always} suffices, which means the fallback procedure $\operatorname{Recover_k}$ is never triggered in this case. 
\end{remark}
\begin{theorem}[Pathwise partial certification]
\label{thm:hidden-two-step-pathwise}
Suppose Assumptions ~\ref{ass:design}--\ref{ass:Penalty} hold.
Run Algorithm~\ref{alg:sequential-newton-cg} with
${\rm is\_perfect}= {\rm false}, J_\star=2$, and $\tau_n = \tau_n^{(2)}$
and 
\begin{align*}
    \sigma_n=2\tau_n^{(2)}\sqrt{2T\log(1/\delta)}/\epsilon,
\end{align*}
with $0<\epsilon \le 1, \delta\in(0,\frac12]$ remain fixed.
For all sufficiently large $n$, the following hold for every fixed $D\in \mathcal{G}_n$ which has probability at least $1-o(1)$:
\begin{enumerate}
    \item $\underset{k\leq T}{\max}\|\widetilde{\beta}_{\bs k}-\beta^*_{\bs k}\|_2\leq\tau_n^{(2)}$ within $J_*=2$ Newton--CG steps.
    \item The published sequence $\{\tbkp\}_{k=1}^T$ satisfies
    $(\epsilon,\delta)$-pathwise partial certification.
    \item For an arbitrary constant $d>0$,
    \begin{equation}\label{eq:path-prd-rate}
     \PP(\operatorname{PRD}_{1:T}
     \leq \sqrt T n^{-3/2}\polylog(n) |\cD)\geq 1-n^{-d}.   
     \end{equation}
\end{enumerate}

\end{theorem}

The complete proofs of Theorems~\ref{thm:hidden-one-step}--\ref{thm:public-three-step}
are collected in Appendix~\ref{app:four-certification-proofs}.

\subsection{Perfect unlearning}\label{ssec: perfect unlearning theory}
\begin{theorem}[Pointwise perfect certification]
\label{thm:public-two-step-pointwise}
Suppose Assumptions~\ref{ass:design}--\ref{ass:Penalty} hold.
Run Algorithm~\ref{alg:sequential-newton-cg} with
${\rm is\_perfect}= {\rm true}, J_\star=2$, $\tau_n=\tau_n^{(2)}$, and
\[
 \sigma_n=2\tau_n^{(2)}\sqrt{2\log(2/\delta)}/\epsilon.
\]
with $0<\epsilon\leq 1, \delta\in(0,\frac12]$ remain fixed. For all $D\in\cG_n$ that has $1-o(1)$ probability, with sufficiently large $n$, the following hold with conditional probability at least $1-n^{-d}$ for an arbitrary constant $d>0$ (over the randomness of Gaussian noises):
\begin{enumerate}
    \item $\max_{k\leq T}\|\widetilde{\beta}_{\bs k}-\beta^*_{\bs k}\|_2\leq\tau_n^{(2)}$ within $J_*=2$ updates for all deletions,
    \item $\{\tbkp\}_{k=1}^T$ satisfies
    $(\epsilon,\delta)$-pointwise perfect certification,
    \item $\operatorname{PRD}_{1:T}=\widetilde O(n^{-3/2})$.
\end{enumerate}
\end{theorem}

\begin{theorem}[Pathwise perfect certification]
\label{thm:public-three-step}\label{thm:perfect-three-step}
Suppose  Assumptions~\ref{ass:design}--\ref{ass:Penalty} hold.
Run Algorithm~\ref{alg:sequential-newton-cg} with
${\rm is\_perfect}= {\rm true}, J_\star=3$, and
\begin{equation}\label{eq:public-tau-choice}
 \tau_n=\frac1n,\qquad
 \sigma_n=\frac{2\sqrt{2T\log(1/\delta)}}{\epsilon n}
\end{equation}
with $0<\epsilon\leq 1, \delta\in(0,\frac12]$ fixed. For all $D\in\cG_n$ that has $1-o(1)$ probability, with sufficiently large $n$, the following hold with conditional probability at least $1-n^{-d}$ for an arbitrary constant $d>0$ (over the randomness of Gaussian noises):
\begin{enumerate}
    \item $\max_{k\leq T}\|\widetilde{\beta}_{\bs k}-\beta^*_{\bs k}\|_2\leq1/n$ within $J_*=3$ updates for all deletions,
    \item $\{\tbkp\}_{k=1}^T$ satisfies
    $(\epsilon,\delta)$-pathwise perfect certification,
    \item $\operatorname{PRD}_{1:T}
 =\widetilde O\left(n^{-3/2}+\frac{\sqrt T}{n^{3/2}}+\frac{T}{n^2}\right).$
\end{enumerate}
\end{theorem}

Comparing Theorem \ref{thm:public-two-step-pointwise} with Theorem \ref{thm:hidden-one-step} (and Theorem \ref{thm:public-three-step} with \ref{thm:hidden-two-step-pathwise}), we  see that, under the prescribed noise levels and center tolerances, perfect unlearning requires one additional Newton--CG iteration compared to partial unlearning. This is because, under perfect unlearning, the $k+1^{\rm th}$ deletion is initialized at the noisy published model $\tbkp$, which brings the publication noise $\sigma_n$ into error accumulation. We might choose a tighter center tolerance to permit a smaller publication-noise variance and thereby control this feedback. Our analysis shows that one extra Newton--CG iteration suffices to attain the corresponding center tolerance with high probability along the deletion sequence.

\paragraph{CG iteration complexity.} For all four settings in Theorems \ref{thm:hidden-one-step}-\ref{thm:perfect-three-step}, CG attains the residual tolerance $\frac{\mu\tau_n}{8}$ within $\polylog n$ iterations per Newton update with high probability. The explicit iteration bound is given in Appendix \ref{app:cg-proof}, Proposition \ref{prop:cg-work}.

\subsection{Sharpness of the maximum Newton steps}
\label{sec:prescribed-update-sharpness}
Theorems \ref{thm:hidden-one-step}--\ref{thm:perfect-three-step} guarantee that, with high probability (when $\cD\in\cG_n$), within a maximum of $J_*=1/2/2/3$ Newton iterations respectively, the desired accuracy $\tau_n$ is achieved and the optimization terminates ($\|g_{\bs k}(u)\|_2\leq \mu\tau_n$) without invoking the fallback procedure $\operatorname{Recover}_k$. 
Theorem \ref{thm:stopping-sharpness} below further shows the aforementioned Newton steps $J_*$ cannot be reduced in the worst case. Note that Theorem \ref{thm:stopping-sharpness} does not include Theorem \ref{thm:hidden-one-step} which requires a single Newton step ($J_*=1$) and it is obviously irreducible: without any Newton updates the model stays at the state of $\sb$ and it loses track for $\sbk$ when deletion progresses.


\begin{theorem}[Sharpness of the maximal Newton Steps]
\label{thm:stopping-sharpness}\label{cor:stopping-sharpness}
Under the conditions of Theorems~\ref{thm:hidden-two-step-pathwise}--\ref{thm:public-three-step}, additionally fix constants
$0<t_-<c'<1$, independently of $n$, and let the
prespecified integer deletion horizon $T=T_n$ satisfy
\[
    \lceil t_-n\rceil \le T \le \lfloor c'n\rfloor
\]
for all sufficiently large $n$.
Let $J_\star=2,2,3$ for pathwise partial, pointwise perfect, and
pathwise perfect unlearning, respectively. Let $N_k$ be the minimum number of Newton iterations such that the gradient gate is met ($\|g_{\bs k}(u)\|\leq \mu\tau_n$).
For each of the three settings, there exists an admissible model family and
events $E_n\subseteq\mathcal G_n$ with $\mathbb P_D(E_n)\to1$ such that,
for every fixed $d>0$ and all sufficiently large $n$, and every $D\in E_n$
\[
 \mathbb P\!\left(
   \max_{1\leq k\leq T}N_k\geq J_\star
   \,\middle|\,D
 \right)
 \geq 1-O(n^{-d}).
\]
\end{theorem}

\section{Numerical Experiments}
\label{sec:simulations}
We examine computational cost, population-risk degradation, and parameter-error trajectories for perfect sequential unlearning with ridge-logistic regression. All Newton--CG results are obtained using Algorithm~\ref{alg:sequential-newton-cg} in the perfect-unlearning setting. All three experiments use the summed loss with penalty $(\lambda/2)\|\beta\|_2^2$, $\lambda=0.1$, and $(\epsilon,\delta)=(1,10^{-6})$. Appendix~\ref{sec:experimental-details} gives the experimental setup.

\subsection{Comparison with GD}
We compare Newton--CG, GD--Armijo, and GD--fixed on spiked Gaussian data ($n=1200$, $p=T=300$, $\rho=1000$). Ten independent clusters pair data, deletion order, and publication innovations across methods. Each method uses its own preceding release and the gate $\|\nabla F_{\backslash k}\|_2\leq\lambda\tau$, with $\tau=1/1200$. Algorithm~\ref{alg:sequential-newton-cg} uses at most two and three Newton updates under pointwise and pathwise calibration, respectively, followed by recovery if needed.

\begin{figure}[!htbp]
\centering
\includegraphics[width=0.58\linewidth]{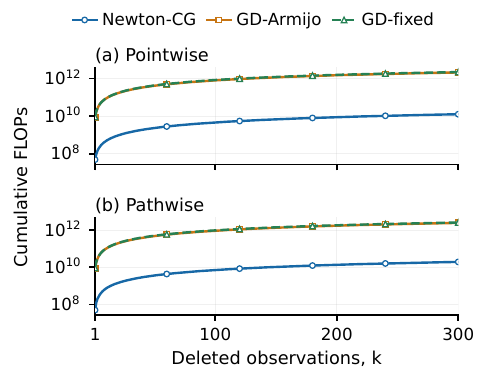}
\caption{Cumulative FLOPs under pointwise (top) and pathwise (bottom) settings.}
\label{fig:perfect-highrho-cumulative}
\end{figure}

Figure~\ref{fig:perfect-highrho-cumulative} shows lower counted costs for Newton--CG. Pathwise paired geometric GD--Armijo/NCG and GD--fixed/NCG FLOP ratios are $125.25$ and $132.06$. All $3000$ deletions completed, including $14$ requiring recovery after the third update; all recovery costs are included. 

\subsection{Population-Risk Degradation}
A separate isotropic study uses $n\in\{600,1200,2400,4800\}$, $p=T=n/4$, $\tau_n=1/n$, and $20$ independent clusters per $n$. Within each calibration, Newton--CG and noisy retraining share publication noise. Relative to numerical noiseless retraining $b_{\mathrm{ref},k}$, we measure
\[
 M_n=\max_{1\leq k\leq T}
 \bigl[R(\tbkp)-R(b_{\mathrm{ref},k})\bigr]_+,
\]
and its matched-retraining analogue, where $R$ is population logistic risk. Each full-path maximum precedes averaging across clusters.

\begin{figure}[!htbp]
\centering
\includegraphics[width=0.58\linewidth]{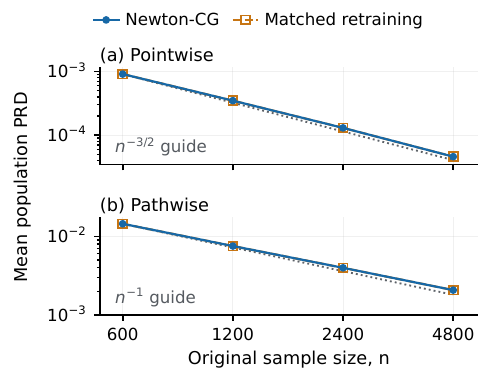}
\caption{Mean full-path population-risk degradation under pointwise (top) and pathwise (bottom) settings with $20$ clusters per $n$. }
\label{fig:prd-n-scaling}
\end{figure}

All $80$ clusters completed. Mean degradation decreases with $n$ and closely tracks noisy retraining (Figure~\ref{fig:prd-n-scaling}). The four-point log--log slopes of the NCG means are $-1.437$ and $-0.933$ for pointwise and pathwise calibration. These descriptive trends do not establish asymptotic rates or separate causal error contributions.

\FloatBarrier
\subsection{Error Propagation}
\label{subsec:error-propagation}
An independent isotropic batch ($n=1200$, $p=T=300$, $\rho=1$) tracks error over $20$ paired data/order/noise clusters. At each deletion, we measure distance to numerical retraining, divided by $\tau=1/1200$, for the entering model, the first complete Newton--CG outer update, and the final accepted center before fresh publication noise.

\begin{figure}[!htbp]
\centering
\includegraphics[width=0.58\linewidth]{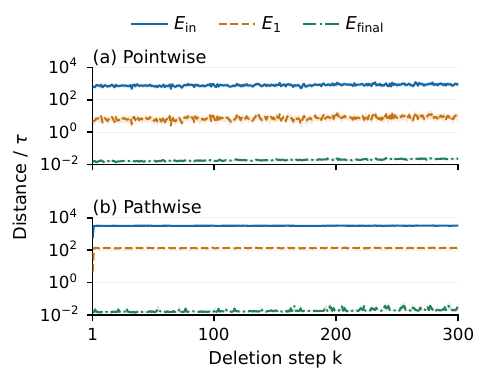}
\caption{Mean distances divided by $\tau$: entering model ($E_{\mathrm{in}}$), first complete update ($E_1$), and final center ($E_{\mathrm{final}}$). Pointwise (top) and pathwise (bottom) panels share a log scale.}
\label{fig:error-propagation}
\end{figure}

All $40$ paths completed $300$ deletions (Figure~\ref{fig:error-propagation}). At $k=300$, pathwise entering and first-update means are $3346.86$ and $139.10$; final means are $0.022$ pointwise and $0.028$ pathwise. Small final errors are partly gate-enforced; natural trajectories do not establish per-round contraction or a causal decomposition.

\section{Concluding Remark}
This paper studies error accumulation in sequential machine unlearning when the parameter dimension $p$ is proportional to the training sample size $n$. Under the stated assumptions, Newton-based updates track exact retraining uniformly over a deletion horizon $T$ proportional to $n$. With the prescribed noise levels $\sigma_n$ and stopping tolerances $\tau_n$, one and two Newton--CG updates per deletion suffice for pointwise and pathwise partial certifiability, respectively, while perfect certifiability requires two and three. These update bounds hold with high probability, and all four settings achieve vanishing population-risk degradation. We further show that the required 1/2/2/3 Newton--CG steps cannot generally be reduced in the worst case. 

Conjugate gradient (CG) reduces the cost of computing Newton updates without explicitly inverting the Hessian. Our analysis controls the additional CG approximation error, and experiments demonstrate substantial computational savings over the tested gradient descent methods in ridge-logistic regression. Together, these results show how inherited approximation error, publication noise, and computation accuracy determine the requirements for sequential unlearning in high dimensions. Extending the analysis to adaptive deletion requests that depend on previously released models and models beyond the convex setting considered here remains an important direction for future work. 

\section*{Acknowledgement}
Arian Maleki would like to thank the National Science Foundation for support under Grants DMS-2515716
and DMS-2210506.

\bibliography{references}

@article{zou2025certified,
  title={Certified Data Removal Under High-dimensional Settings},
  author={Zou, Haolin and Auddy, Arnab and Kwon, Yongchan and Rad, Kamiar Rahnama and Maleki, Arian},
  journal={arXiv preprint arXiv:2505.07640},
  year={2025}
}

@inproceedings{neel2021descent,
  title={Descent-to-delete: Gradient-based methods for machine unlearning},
  author={Neel, Seth and Roth, Aaron and Sharifi-Malvajerdi, Saeed},
  booktitle={Algorithmic Learning Theory},
  pages={931--962},
  year={2021},
  organization={PMLR}
}

@article{guo2019certified,
  title={Certified data removal from machine learning models},
  author={Guo, Chuan and Goldstein, Tom and Hannun, Awni and Van Der Maaten, Laurens},
  journal={arXiv preprint arXiv:1911.03030},
  year={2019}
}

@inproceedings{cao2015towards,
  title={Towards making systems forget with machine unlearning},
  author={Cao, Yinzhi and Yang, Junfeng},
  booktitle={2015 IEEE symposium on security and privacy},
  pages={463--480},
  year={2015},
  organization={IEEE}
}

@article{ginart2019making,
  title={Making ai forget you: Data deletion in machine learning},
  author={Ginart, Antonio and Guan, Melody and Valiant, Gregory and Zou, James Y},
  journal={Advances in neural information processing systems},
  volume={32},
  year={2019}
}

@article{sekhari2021remember,
  title={Remember what you want to forget: Algorithms for machine unlearning},
  author={Sekhari, Ayush and Acharya, Jayadev and Kamath, Gautam and Suresh, Ananda Theertha},
  journal={Advances in Neural Information Processing Systems},
  volume={34},
  pages={18075--18086},
  year={2021}
}

@article{gupta2021adaptive,
  title={Adaptive machine unlearning},
  author={Gupta, Varun and Jung, Christopher and Neel, Seth and Roth, Aaron and Sharifi-Malvajerdi, Saeed and Waites, Chris},
  journal={Advances in Neural Information Processing Systems},
  volume={34},
  pages={16319--16330},
  year={2021}
}

@inproceedings{bourtoule2021machine,
  title={Machine unlearning},
  author={Bourtoule, Lucas and Chandrasekaran, Varun and Choquette-Choo, Christopher A and Jia, Hengrui and Travers, Adelin and Zhang, Baiwu and Lie, David and Papernot, Nicolas},
  booktitle={2021 IEEE Symposium on Security and Privacy (SP)},
  pages={141--159},
  year={2021},
  organization={IEEE}
}

@article{suriyakumar2022algorithms,
  title={Algorithms that approximate data removal: New results and limitations},
  author={Suriyakumar, Vinith and Wilson, Ashia C},
  journal={Advances in Neural Information Processing Systems},
  volume={35},
  pages={18892--18903},
  year={2022}
}

@inproceedings{qiao2025hessian,
  title={Hessian-free online certified unlearning},
  author={Qiao, Xinbao and Zhang, Meng and Tang, Ming and Wei, Ermin},
  booktitle={International Conference on Learning Representations},
  volume={2025},
  pages={32675--32711},
  year={2025}
}

@inproceedings{
koloskova2025certified,
title={Certified Unlearning for Neural Networks},
author={Anastasia Koloskova and Youssef Allouah and Animesh Jha and Rachid Guerraoui and Sanmi Koyejo},
booktitle={Forty-second International Conference on Machine Learning},
year={2025}
}

@inproceedings{
bui2026how,
title={How to Cure Newton for Unlearning Neural Networks? An Empirical Study from the Hessian Perspective},
author={Nhung Bui and Xinyang Lu and Rachael Hwee Ling Sim and See-Kiong Ng and Bryan Kian Hsiang Low},
booktitle={The Fourteenth International Conference on Learning Representations},
year={2026},
url={https://openreview.net/forum?id=dHz2LBCyTh}
}

@article{kairouz2013TheCT,
  title={The Composition Theorem for Differential Privacy},
  author={Peter Kairouz and Sewoong Oh and Pramod Viswanath},
  journal={IEEE Transactions on Information Theory},
  year={2013},
  volume={63},
  pages={4037-4049}
}

@inproceedings{bun2016concentrated,
  author    = {Mark Bun and Thomas Steinke},
  title     = {Concentrated Differential Privacy: Simplifications,
               Extensions, and Lower Bounds},
  booktitle = {Theory of Cryptography -- {TCC 2016-B}},
  series    = {Lecture Notes in Computer Science},
  volume    = {9985},
  pages     = {635--658},
  publisher = {Springer},
  year      = {2016},
  doi       = {10.1007/978-3-662-53641-4_24}
}

@article{pandey2025gaussian,
  title   = {{Gaussian} Certified Unlearning in High Dimensions:
             A Hypothesis Testing Approach},
  author  = {Pandey, Aaradhya and Auddy, Arnab and Zou, Haolin
             and Maleki, Arian and Kulkarni, Sanjeev},
  journal = {arXiv preprint arXiv:2510.13094},
  year    = {2025},
  doi     = {10.48550/arXiv.2510.13094},
  url     = {https://arxiv.org/abs/2510.13094}
}

@article{hestenes1952methods,
  title={Methods of conjugate gradients for solving linear systems},
  author={Hestenes, Magnus R and Stiefel, Eduard and others},
  journal={Journal of research of the National Bureau of Standards},
  volume={49},
  number={6},
  pages={409--436},
  year={1952}
}
\clearpage
\appendix
\section*{Supplementary Materials}

\section{Assumptions}\label{sec: assumptions}
\begin{assumption}[Gaussian elliptical design]\label{ass:design}
The training pairs $(x_i,y_i)$ are independent draws from a common population. Their covariates have the representation
\[
    x_i=\frac{1}{\sqrt p}\Sigma^{1/2}\xi_i,
    \qquad \xi_i\stackrel{\mathrm{iid}}{\sim}N(0,I_p),
\]
where the spectrum of $\Sigma$ is uniformly bounded:
\[
    0<c_\Sigma\leq \lambda_{\min}(\Sigma)
    \leq \lambda_{\max}(\Sigma)\leq C_\Sigma<\infty .
\]
The constants $c_{\Sigma}, C_{\Sigma}$ are independent of $n,p$. The components $x_i$ and $y_i$ within a pair need not be independent. 
\end{assumption}

\begin{assumption}[Response tails]\label{ass:response}
There exists $C_y(n)=O((\log n)^{C_y^0})$ and a tail probability $q_y(n)$ such that
\[
    \mathbb P(|y_i|>C_y(n))\leq q_y(n),
    \qquad
    nq_y(n)=o(1),
    \qquad
    \mathbb E |y_i|^{2s}\leq C_{y,s}<\infty .
\]

\end{assumption}

\begin{assumption}[Loss]\label{ass:loss}

    For every $y$, the loss $\ell(y,\cdot)$ is nonnegative, proper convex and $C^3$. For fixed constants $C,s >0$, and all $(y,t)$,
    \begin{align*}
        \max\{\ell(y,t),|\partial_t\ell(y,t)|,|\partial_t^3\ell(y,t)|\} \leq C(1+|y|^s+|t|^s).
    \end{align*}
\end{assumption}
\begin{assumption}[Penalty]\label{ass:Penalty}
The regularizer $r:\mathbb R^p\to\mathbb R_+$ is $C^2$,
proper convex, and $\nu$-strongly convex for fixed $\nu>0$.
It is separable, $r(\beta)=\sum_{j=1}^p r_j(\beta_j)$,
and is normalized by
\[
  r(0)=0,\qquad \nabla r(0)=0.
\]
Consequently,
\[
  r(\beta)\ge\frac{\nu}{2}\|\beta\|_2^2,
  \qquad \beta\in\mathbb R^p.
\]
This normalization gives the a priori radius bound for the regularized
empirical minimizer; the origin is the reference minimizer of $r$.
The Hessian has the global bounds
\[
\begin{aligned}
  \sup_{\beta\in\mathbb R^p}
     \|\nabla^2 r(\beta)\|_{\mathrm{op}}&\le B_r(n),\\
  \|\nabla^2 r(\beta)-\nabla^2 r(\beta')\|_{\mathrm F}
    &\le C_{rr}(n)\|\beta-\beta'\|_2,
\end{aligned}
\]
where the second inequality holds for all
$\beta,\beta'\in\mathbb R^p$,
$C_{rr}(n)=O((\log n)^{C_{rr}^0})$ is deterministic with fixed
$C_{rr}^0\ge0$, and $B_r(n)$ is a deterministic polylogarithmic bound.
\end{assumption}

Gaussianity is used for conditional leave-one-out linear forms, the
design operator bound, and the test-population moment calculation.
An extension beyond Gaussian covariates would require corresponding
conditional concentration and test-moment bounds to be proved. No such
extension is asserted here.

\section{Shared tools for tracking exact retraining}
\label{app:tracking}
\paragraph{Notation for the proofs.}
For fixed data and a specified algorithmic configuration, let $C_k^H$
be the center in Algorithm~\ref{alg:sequential-newton-cg} with 
${\rm is\_perfect}= {\rm false}$ and
$C_k^P(h)$ the center when 
${\rm is\_perfect}= {\rm true}$ at the numerical past history $h$.
Thus $C_k^H=\widetilde{\beta}_{\bs k}$ for the hidden algorithm, while
$C_k^P(Y_{<k}^P)=\widetilde{\beta}_{\bs k}$ along the actual public path.
The actual releases are $Y_k^H=C_k^H+Z_k$ and
$Y_k^P=C_k^P(Y_{<k}^P)+Z_k$. The chronological retraining reference is
$V_k=\beta^*_{\bs k}+Z'_k$, where the $Z'_k$ are mutually independent
$N(0,\sigma_n^2I_p)$ draws, also independent of mechanism noise.
Each theorem uses its displayed noise variance.

For the guarded algorithms, write $B_{\mathrm N}=J_\star$ for the
maximum Newton--CG budget in Algorithm~\ref{alg:sequential-newton-cg}:
$J_\star=1,2,2,3$ for pointwise partial, pathwise partial, pointwise
perfect, and pathwise perfect unlearning, respectively. The gradient
gate is checked after every update, including the last allowed update.
Recovery is invoked only if that last gate fails. The tracking claims
below establish gate acceptance within this budget without recovery;
they do not merely bound the number of attempted Newton updates.

All norms without subscripts are Euclidean. Write
$b_k=\beta^*_{\bs k}$, $X_k=X_{\bs k}$, $H_k=H_{\bs k}$ and
$g_k=g_{\bs k}$ in this appendix; $b_k$ is an exact fit, whereas $z_k$
in Section~2.1 denotes the observation deleted at round $k$.
The order is fixed independently of the complete observation pairs.
All constants below are deterministic. Fix $d>0$ once, put
$b=n^{-d}$, and compute envelopes for $K=\lfloor c'n\rfloor$.
Take $n$ large enough that $K\geq1$ and
$0<\gamma_-\leq p/n\leq\gamma_+<\infty$ for fixed constants.
Later accuracy exponents may differ from $d$.

This appendix collects the shared tracking tools: the data-only event
in Appendix~\ref{app:data-event}, deletion-direction and local Hessian
estimates, the Newton--CG recursions and direct gradient bound in
Appendix~\ref{app:newton-cg-recursion}, and the common tolerances and
original Gaussian innovations below. The complete proofs of
Theorems~\ref{thm:hidden-one-step}--\ref{thm:public-three-step} are in
Appendix~\ref{app:four-certification-proofs}, one proof per theorem with
tracking, certification, and population-risk degradation together.
The hidden-state arguments in Appendices~\ref{app:proof-hidden-one-step}
and~\ref{app:proof-hidden-two-step} use no publication noise; the public
arguments in Appendices~\ref{app:proof-public-two-step}
and~\ref{app:proof-public-three-step} use the original-innovation event
before tracking is considered. Appendices~\ref{app:pathwise-proof},
\ref{app:prd-proof}, \ref{app:cg-proof}, and~\ref{app:recovery-proof}
provide the shared certification, risk, CG, and recovery tools.
The stopping-count theorem has its proof in
Appendix~\ref{app:stopping-sharpness}, after its supporting constructions.

\begin{lemma}[Existence and uniform strong convexity]
\label{lem:uniform_strong_convexity}
Every retained or additional leave-one-out objective has a unique
minimizer. For every dataset, every $k$, and every parameter $u$,
$H_k(u)\succeq\mu I_p$ and
$\|u-b_k\|\leq\|g_k(u)\|/\mu$.
\end{lemma}
\begin{proof}
The objective is continuous and at least $\mu\|u\|^2/2$, hence coercive.
Its loss Hessian is positive semidefinite and the penalty contributes
at least $\lambda\nu I_p$. Strong convexity gives uniqueness. Integrating
the Hessian from $b_k$ to $u$ gives
$\mu\|u-b_k\|^2\leq\langle g_k(u),u-b_k\rangle
\leq\|g_k(u)\|\|u-b_k\|$.
\end{proof}

\subsection{A measurable common data event}
\label{app:data-event}
Define the following thresholds from the primitive model constants:
\begin{align}
 B^2&=2Cn(1+C_y(n)^s)/\mu,
 &m_1&=n(K+1),&m_2&=K(n+p),\notag\\
 M^2&=\frac{8C_\Sigma}{p}\{(n+p)\log9+\log(8/b)\},
 &B_X&=\max\{1,M\},\label{eq:data-thresholds-a}\\
 t&=B\sqrt{\frac{2C_\Sigma}{p}\log\frac{8m_1}{b}},
 &h&=\frac{B_X}{\mu}\sqrt{\frac{2C_\Sigma}{p}\log\frac{8m_2}{b}}.
 \label{eq:data-thresholds-b}
\end{align}
Here $M=O(1)$, $B/\sqrt p$ and $t$ are polylogarithmic, and
$h=O(\sqrt{\log(en)/p})$. Let $b_k^{(-i)}$ be the exact fit after
also removing observation $i$ if it is retained, and $b_k$ otherwise.
Define $\mathcal G_n$ by the simultaneous data properties
\begin{equation}\label{eq:data-event-definition}
 \begin{gathered}
 \max_i|y_i|\leq C_y(n),\qquad \|X\|_{\mathrm{op}}\leq M,\\
 \max_{0\leq k\leq K,\,1\leq i\leq n}|x_i^\top b_k^{(-i)}|\leq t,
 \end{gathered}
\end{equation}
\begin{equation}\label{eq::independence concentration}
 \max_{1\leq k\leq K}
 \max\{\|H_k(b_k)^{-1}x_k\|_\infty,
          \|X_kH_k(b_k)^{-1}x_k\|_\infty\}\leq h.
\end{equation}
This event contains only data functions, not an approximate iterate or
a success predicate. As implicit in the statistical model, losses and
derivatives are observation-measurable. The exact fit is measurable:
enumerate rational vectors and select the first whose objective is
within $1/m$ of its rational infimum. Continuity identifies that
infimum with the minimum, and strong convexity forces these selections
to converge to the unique minimizer. Hessian entries and their inverses
are then measurable. Efficient recognition of this event is not
required and is not a screening step of the algorithm.

\begin{lemma}[Full-prefix data probability and score envelopes]
\label{lem:uniform-score-bound}\label{lem:common-data-event}
The event in \eqref{eq:data-event-definition}--\eqref{eq::independence concentration}
satisfies
\begin{equation}\label{eq:data-failure-bound}
 \PP_D(\mathcal G_n^c)\leq nq_y(n)+3b/4\leq p_n=o(1).
\end{equation}
Define, in the displayed order,
\begin{equation}\label{eq:score-envelopes}
 \begin{split}
 S_0&=C(1+C_y^s+t^s),\qquad P_0=t+M^2S_0/\mu,\\
 S&=C(1+C_y^s+P_0^s),\qquad D_0=MS/\mu.
 \end{split}
\end{equation}
On $\mathcal G_n$, every $0\leq k\leq K$ and $1\leq i\leq n$ obey
\begin{equation}\label{eq:exact-fit-envelopes}
 \begin{split}
 \|b_k\|&\leq B,\qquad |x_i^\top b_k|\leq P_0,\\
 |\dot\ell(y_i,x_i^\top b_k)|,\ |\dddot\ell(y_i,x_i^\top b_k)|&\leq S,
 \qquad \|\nabla\ell_i(b_k)\|\leq MS.
 \end{split}
\end{equation}
In particular $\|b_k-b_{k-1}\|\leq D_0$ for $1\leq k\leq K$.
The same event and envelopes serve every $T\leq K$.
\end{lemma}
\begin{proof}
The response event fails with probability at most $nq_y(n)$.
For the design, take $1/4$-nets of the unit spheres in $\RR^n$ and
$\RR^p$, of cardinalities at most $9^n$ and $9^p$. Such nets follow
from a maximal separated set and disjoint radius-$1/8$ balls inside
a radius-$9/8$ ball. Approximating the two maximizing unit vectors
one at a time shows that the operator norm is at most twice the
largest absolute bilinear form over these nets. Each fixed form
has Gaussian variance at most $C_\Sigma/p$. Its two-sided tail and
a union bound give
\[
 \PP(\|X\|_{\mathrm{op}}>M)
 \leq2\,9^{n+p}\exp\{-pM^2/(8C_\Sigma)\}=b/4.
\]
All row and retained-submatrix norms are then at most $M$.

Objective comparison at zero, nonnegative loss and
$r(u)\geq\nu\|u\|^2/2$ give, on the response event,
\begin{equation}\label{eq:loo-radius-bound}
 \|b_k\|^2,\ \|b_k^{(-i)}\|^2
 \leq\frac{2}{\mu}\sum_{j=1}^n\ell(y_j,0)\leq B^2.
\end{equation}
For fixed $k,i$, condition on all pairs except $(x_i,y_i)$.
The fit $b_k^{(-i)}$ is fixed and independent of $x_i$, including
when $i$ is already removed. On its radius restriction the predictor
has conditional variance at most $C_\Sigma B^2/p$. Consequently,
after integrating that conditional estimate,
\[
 \PP\{|x_i^\top b_k^{(-i)}|>t,\ \|b_k^{(-i)}\|\leq B\}
 \leq2\exp\{-pt^2/(2C_\Sigma B^2)\}=b/(4m_1).
\]
The union over $m_1$ fits costs at most $b/4$ on the response event.
This calculation does not condition a held-out covariate on its own
response or on the full response event.

For the last property, condition at deletion $k$ on the retained
pairs. The retained $H_k(b_k)$ and $X_k$ are independent of $x_k$.
The parameter-coordinate variances are at most
$C_\Sigma/(p\mu^2)$; the retained-predictor variances are at most
$C_\Sigma M^2/(p\mu^2)$ on $\|X_k\|_{\mathrm{op}}\leq M$.
Each scalar failure at threshold $h$ is thus at most $b/(4m_2)$.
There are at most $m_2=K(n+p)$ coordinates across the whole path.
Integrating the conditional estimate and only then intersecting the
design event gives an additional $b/4$. Adding these three allocations
and the response failure proves \eqref{eq:data-failure-bound}.

If $i$ is retained, its addition to the leave-one-out objective gives
$\|g_k(b_k^{(-i)})\|\leq MS_0$. Strong convexity therefore implies
$\|b_k-b_k^{(-i)}\|\leq MS_0/\mu$ and
$|x_i^\top b_k|\leq t+M^2S_0/\mu=P_0$.
For removed $i$ the two fits coincide. Loss growth proves the remaining
envelopes. Finally $g_k(b_{k-1})=-\dot\ell(y_k,x_k^\top b_{k-1})x_k$,
so the same strong-convexity inequality gives the shift bound.
\end{proof}
Only $nq_y(n)=o(1)$ is assumed; a polynomial rate for all of $p_n$
does not follow. One may put $\mathcal G_{n,T}=\mathcal G_n$ for every
$T$. For an independently randomized order, first condition on that
order and use the same bound; the event then depends on data and order.
Nothing asserts simultaneous coverage of every permutation or adaptive
deletion rule.

\subsection{Deletion error and local Hessian bounds}
Define
\begin{equation}\label{eq:oracle-constants}
 \begin{split}
 A_0&=M^2S+\lambda C_{rr},\qquad
 c_n=2SA_0D_0h/\mu,\\
 \rho_n&=Sh(1+B_XA_0D_0/\mu),\qquad
 R_n=D_0+2+\log(en).
 \end{split}
\end{equation}
The radius is chosen before the local derivative constants and
tolerances. It leaves logarithmic slack for public-noise norms at
fixed privacy parameters. Set
\begin{align}
 V&=C\{1+C_y^s+(P_0+MR_n)^s\},\notag\\
 U&=3C\{1+C_y^s+(P_0+MR_n+1)^s\},\notag\\
 L&=M^2U+\lambda B_r,\qquad
 L_H=M^3V+\lambda C_{rr},\qquad A=M^2V+\lambda C_{rr},
 \label{eq:local-envelope-definitions}\\
 a_n&=A\rho_n/\mu,\qquad b_n=L_H/(2\mu).
 \label{eq:recursion-constants}
\end{align}
All envelopes have fixed polylogarithmic growth; specifically
$c_n,\rho_n,a_n=\widetilde O(n^{-1/2})$.

\begin{lemma}[A deletion correction and its direction]
\label{lem:Oracle Shift Bound}\label{lemma:: One-step oracle error}
\label{lem:residual_gradient}
On $\mathcal G_n$, for $v=b_{k-1}$ and $z=b_k$,
\begin{align}
 \|v-z\|&\leq D_0,
 &\|\mathcal N_k(v)-z\|&\leq c_n,
 \label{eq:oracle-newton-rewrite}\\
 \max\{\|H_k(v)^{-1}g_k(v)\|_\infty,
     \|X_kH_k(v)^{-1}g_k(v)\|_\infty\}&\leq\rho_n.
 \label{eq:rho-k-definition}
\end{align}
\end{lemma}
\begin{proof}
Write $H=H_k$, $g=g_k$ and
$s_k=\dot\ell(y_k,x_k^\top v)$, so $g(v)=-s_kx_k$ and $|s_k|\leq S$.
Predictors on $[v,z]$ are bounded by $P_0$. For $a,b$ on this segment,
a loss Hessian difference is $X_k^\top D_\ell X_k$ with diagonal
$D_\ell$ satisfying $\|D_\ell\|_{\mathrm F}\leq SM\|a-b\|$.
The penalty difference is diagonal with Frobenius norm at most
$\lambda C_{rr}\|a-b\|$. For $w=H(z)^{-1}x_k$,
\eqref{eq::independence concentration} and
$\|Dw\|\leq\|D\|_{\mathrm F}\|w\|_\infty$ for diagonal $D$ give
\[
 \|[H(a)-H(b)]w\|\leq A_0\|a-b\|h.
\]
Let $\bar H=\int_0^1 H(v+t(z-v))\,dt$. Optimality implies
$z-v=\bar H^{-1}s_kx_k$ and
$\mathcal N_k(v)-z=[H(v)^{-1}-\bar H^{-1}]s_kx_k$.
The ordered identity
\begin{align*}
 H(v)^{-1}-\bar H^{-1}
 &=H(v)^{-1}[H(z)-H(v)]H(z)^{-1}\\
 &\quad+\bar H^{-1}[\bar H-H(z)]H(z)^{-1}
\end{align*}
requires no commutation. Each inverse has norm at most $\mu^{-1}$,
and each action costs at most $A_0D_0h$. This proves the $c_n$ bound.
The further identity
\[
 H(v)^{-1}x_k=H(z)^{-1}x_k+
 H(v)^{-1}[H(z)-H(v)]H(z)^{-1}x_k
\]
proves \eqref{eq:rho-k-definition}, including the $B_X$ factor for
multiplication by $X_k$. Separability, not a quadratic penalty or
diagonal covariance, supplies the diagonal regularizer action.
\end{proof}

\begin{lemma}[Local Hessian control]
\label{lem:lipschitz_hessian}
On $\mathcal G_n$, every $a,b\in B(b_k,R_n)$ satisfy
\begin{align}
 \mu I_p\preceq H_k(a)&\preceq LI_p,
 \label{eq:local-spectrum-rewrite}\\
 \|H_k(a)-H_k(b)\|_{\mathrm{op}}&\leq L_H\|a-b\|,
 \label{eq:local-lipschitz-rewrite}\\
 \|H_k(a)-H_k(b)\|_{\mathrm{op}}
 &\leq A\max\{\|X_k(a-b)\|_\infty,\|a-b\|_\infty\}.
 \label{eq:local-directional-rewrite}
\end{align}
\end{lemma}
\begin{proof}
Every predictor in the ball is bounded by $P_0+MR_n$. The identity
\[
 \ddot\ell(y,t)=\dot\ell(y,t+1)-\dot\ell(y,t)
   -\int_0^1(1-u)\dddot\ell(y,t+u)\,du
\]
bounds its second derivative by $U$: the common bracket costs at most
$2C+C/2\leq3C$. Convexity makes that derivative nonnegative.
The design and penalty upper bounds give $L$. The loss contribution
to a Hessian difference is bounded by
$M^2V\|X_k(a-b)\|_\infty\leq M^3V\|a-b\|$.
For the penalty, apply the global Frobenius Lipschitz condition to
vectors differing only in coordinate $j$ to get
$|r_j''(a_j)-r_j''(b_j)|\leq C_{rr}|a_j-b_j|$.
Its maximum and its Euclidean bound prove the other two assertions.
These inequalities hold for every input in the ball, including
data-dependent and noise-dependent inputs.
\end{proof}

\subsection{Newton--CG recursions and the actual gradient gate}
\label{app:newton-cg-recursion}
\begin{lemma}[Refined and quadratic error bounds]
\label{lem:accum_err}
On $\mathcal G_n$, let $v=b_{k-1}$, $z=b_k$, $u\in B(z,R_n)$ and
$\Delta=u-v$. Define
$m_k(\Delta)=\max\{\|X_k\Delta\|_\infty,\|\Delta\|_\infty\}$.
Then
\begin{align}
 \|\mathcal N_k(u)-z\|
 &\leq c_n+a_n\|\Delta\|+b_n\|\Delta\|^2,
 \label{eq:refined-hidden-newton-rewrite}\\
 \|\mathcal N_k(u)-z\|
 &\leq c_n+(A/\mu)m_k(\Delta)(\|\Delta\|/2+D_0),
 \label{eq:refined-public-newton-rewrite}\\
 \|\mathcal N_k(u)-z\|&\leq b_n\|u-z\|^2.
 \label{eq:quadratic-newton-rewrite}
\end{align}
For a CG direction with true residual $r=g_k(u)-H_k(u)d$, each bound
gains at most $\|r\|/\mu\leq\tau_n/8$. If also $u-d$ belongs to
the ball, the actual post-update gradient satisfies
\begin{equation}\label{eq:direct-gradient-gate}
 \|g_k(u-d)\|\leq\|r\|+(L_H/2)\|d\|^2.
\end{equation}
\end{lemma}
\begin{proof}
Abbreviate $H=H_k$, $g=g_k$. Both $v$ and $u$ belong to the convex
ball. Direct subtraction gives
\begin{align*}
 \mathcal N_k(u)-\mathcal N_k(v)
 &=H(u)^{-1}\int_0^1[H(u)-H(v+t\Delta)]\Delta\,dt\\
 &\quad+H(u)^{-1}[H(u)-H(v)]H(v)^{-1}g(v).
\end{align*}
The first term is at most $b_n\|\Delta\|^2$. In the second, the
diagonal loss/penalty decomposition used above and
\eqref{eq:rho-k-definition} give $A\rho_n\|\Delta\|/\mu$.
This proves \eqref{eq:refined-hidden-newton-rewrite}.
Alternatively \eqref{eq:local-directional-rewrite} bounds the first
term by $(A/(2\mu))m_k(\Delta)\|\Delta\|$, and the second by
$(A/\mu)m_k(\Delta)D_0$, since
$\|H(v)^{-1}g(v)\|\leq MS/\mu=D_0$.
This proves \eqref{eq:refined-public-newton-rewrite}.
Taylor integration along $[z,u]$ gives
\eqref{eq:quadratic-newton-rewrite}.
The exact residual identity $u-d=\mathcal N_k(u)+H(u)^{-1}r$
proves the CG versions.
Finally, when its segment is in the ball,
\[
 g(u-d)=r+\int_0^1[H(u)-H(u-td)]d\,dt.
\]
Integrating the Lipschitz bound proves
\eqref{eq:direct-gradient-gate}. We first place the output in the ball
using its distance estimate and only then apply this gradient identity.
\end{proof}

\subsection{Deterministic center tolerances}
\label{app:hidden-proof}
Write $\kappa_n=L/\mu\geq1$.
Define
\begin{equation}\label{eq:two-step-center-tolerance}
 \begin{gathered}
 \tau_n^{(1)}=4\kappa_n c_n+n^{-1/2},\qquad
 w_n=c_n+n^{-1/2},\\
 \tau_n^{(2)}=(1+16L_H/\mu)w_n^2.
 \end{gathered}
\end{equation}
$\tau_n^{(1)}$ and $\tau_n^{(2)}$ are the same quantities used in Theorems \ref{thm:hidden-one-step}, \ref{thm:hidden-two-step-pathwise} and \ref{thm:public-two-step-pointwise}.

\subsection{Original Gaussian innovations}
\label{app:three-step-proof}
For fixed $D\in\mathcal G_n$, any $\rho\in(0,1)$ and the original
independent $Z_1,\ldots,Z_T\sim N(0,\sigma_n^2I_p)$, let
\begin{align}
 U_N&=\sigma_n\{\sqrt p+\sqrt{2\log(2T/\rho)}\},
 \label{eq:noise-norm-rewrite}\\
 U_D&=B_X\sigma_n\sqrt{2\log(4T(n+p)/\rho)}.
 \label{eq:noise-directions-rewrite}
\end{align}
With conditional probability at least $1-\rho$, simultaneously
\begin{equation}\label{eq:innovation-event}
 \max_{j\leq T}\|Z_j\|\leq U_N,\qquad
 \max_{j\leq T}\max\{\|Z_j\|_\infty,\|XZ_j\|_\infty\}\leq U_D.
\end{equation}
To verify the norm tail directly, for a standard Gaussian vector $G$
and $W=\|G\|^2$,
$\log\EE e^{s(W-p)}\leq ps^2/(1-2s)$ for $0<s<1/2$.
Chernoff at $s=\sqrt t/(\sqrt p+2\sqrt t)$ gives
$\PP(W\geq p+2\sqrt{pt}+2t)\leq e^{-t}$.
The larger threshold $(\sqrt p+\sqrt{2t})^2$ gives the stated norm
bound with a union cost $\rho/2$. The scalar Gaussian tail and row
norm bound $M$ give cost $\rho/2$ for the at most $T(n+p)$ two-sided
coordinate/predictor tests. No Gaussian law has been conditioned on
future success. Write $Z_0=0$ only as deterministic notation; no initial
noise is drawn or released.

All sufficiently-large-$n$ thresholds in the tracking arguments in
Appendix~\ref{app:four-certification-proofs} are uniform over a
prespecified $T\leq K$. The data event covers all prefixes
simultaneously; separately rerunning every horizon-dependent noise
calibration does not receive a shared $\rho$ budget from this statement.

\section{Certification tools and proofs of the main theorems}
\label{app:pathwise-proof}
All probabilities in this section are conditional on a fixed dataset
and order. We use the release processes and independent chronological
retraining reference defined at the start of Appendix~\ref{app:tracking}.
Pointwise results compare each actual marginal with $\mathcal L(V_k)$;
pathwise results compare the joint transcript with $\mathcal L(V_{1:T})$.
The public marginal is generally a Gaussian mixture. We compare public
kernels at the same numerical past history, rather than conditioning
on the full random center sequence.

\begin{theorem}[Gaussian conditional-kernel comparison]
\label{thm:gaussian-certification}\label{thm:pathwise-certification}
Let a comparison process use measurable centers $m_k(h)$ satisfying
$\|m_k(h)-b_k\|\leq\tau$ at every history $h$, followed by fresh
$N(0,\sigma^2I_p)$ noise. Let $Q$ be the joint law of independently
noised exact fits $b_1,\ldots,b_T$. For $\delta_G\in(0,1)$ the two
joint laws obey both $(\epsilon,\delta_G)$ inequalities whenever
\begin{equation}\label{eq:gaussian-path-calibration}
 v/2+\sqrt{2v\log(1/\delta_G)}\leq\epsilon,
 \qquad v=T\tau^2/\sigma^2.
\end{equation}
For a single release, the same conclusion holds with $T=1$, including
a mixture of bounded centers whose fresh noise is independent of them.
\end{theorem}
\begin{proof}
At each common history write the two Gaussian conditional densities as
$p_k(\cdot\mid h)$ and $q_k(\cdot\mid h)$. Completing the square gives,
for every $\alpha>1$,
\begin{align*}
 \int p_k(x\mid h)^\alpha q_k(x\mid h)^{1-\alpha}\,dx
 &=\exp\left\{\frac{\alpha(\alpha-1)
                   \|m_k(h)-b_k\|^2}{2\sigma^2}\right\}\\
 &\leq\exp\{\alpha(\alpha-1)\tau^2/(2\sigma^2)\}.
\end{align*}
The same identity holds with $p_k,q_k$ interchanged. Factoring joint
densities chronologically and integrating the last coordinate first,
then repeating, gives both directional moment bounds
$\EE\exp\{t\mathcal L\}\leq\exp\{t(t+1)v/2\}$, $t>0$.
Here $\mathcal L$ is the appropriate log likelihood ratio and each
expectation is under its numerator law. Equivalently
$\EE e^{t(\mathcal L-v/2)}\leq e^{t^2v/2}$.
If $v=0$, the laws coincide. Otherwise Chernoff at
$t=\sqrt{2\log(1/\delta_G)/v}$ bounds by $\delta_G$ the numerator
probability that $\mathcal L>v/2+\sqrt{2v\log(1/\delta_G)}$.
For any measurable event, split the integral into this set and its
complement. On the complement the numerator density is at most
$e^\epsilon$ times the denominator density. This proves one
distributional inequality, and the reverse moment proves the other.
This is a conditional-kernel calculation, also consistent with the
Gaussian divergence framework of \citet{bun2016concentrated}; it does
not compose marginal guarantees.

For a single random center $C$ supported on $\|C-b_k\|\leq\tau$,
apply the $T=1$ result at every fixed center and integrate both
inequalities against its law. In the reverse direction $Q_k(A)$ is
constant under that integral. This proves the mixture statement and
applies to a public marginal regardless of its preceding history law.
\end{proof}

\begin{lemma}[Transfer for a capped reference process]
\label{lem:actual-law-transfer}
Within this lemma, $C_k^P$ denotes the center of a public capped
reference process. Suppose it has
\[
 \PP_D\{\exists k\leq T:
       \|C_k^P(Y_{<k}^P)-b_k\|>\tau\}\leq\rho.
\]
There is a comparison law $\bar P_D$ with all-history center bound
$\tau$ and $\operatorname{TV}(P_D,\bar P_D)\leq\rho$. If
$\bar P_D$ and the appropriate retraining reference $Q_D$ are
bilaterally $(\epsilon,\delta_G)$-indistinguishable, then every
measurable event satisfies
\begin{align}
 P_D(A)&\leq e^\epsilon Q_D(A)+\delta_G+\rho,
 \label{eq:actual-forward-transfer}\\
 Q_D(A)&\leq e^\epsilon P_D(A)+\delta_G+e^\epsilon\rho.
 \label{eq:actual-reverse-transfer}
\end{align}
The same transfer holds for each projected marginal.
\end{lemma}
\begin{proof}
At common history $h$, let $f_k(h)=C_k^P(h)$ be the output of the
actual capped computation, with its actual residual and early-stop
rules. Define solely for this proof
\[
 m_k(h)=
 \begin{cases}
 f_k(h),&\|f_k(h)-b_k\|\leq\tau,\\
 b_k,&\text{otherwise}.
 \end{cases}
\]
Its Gaussian release kernels define $\bar P_D$.
The capped map is measurable: its arithmetic operations and comparison
branches are measurable, CG terminates in at most $p$ exact-arithmetic
steps on an SPD matrix, and there are at most $J$ outer updates, with
the stated zero-gradient convention. Measurability of $b_k$ was shown
in Appendix~\ref{app:data-event}.

Couple the actual and comparison processes with identical original
independent Gaussian innovations. They agree until the first actual
unsafe center: before that time their histories and candidate centers
coincide. Hence the probability of unequal transcripts is at most
$\rho$ under the actual law, which bounds total variation. No second
tracking estimate under the reference law is needed. Projection can
only decrease this total variation. Finally,
$P_D(A)\leq\bar P_D(A)+\rho$ gives
\eqref{eq:actual-forward-transfer}, whereas substituting
$\bar P_D(A)\leq P_D(A)+\rho$ into the reverse comparison gives
\eqref{eq:actual-reverse-transfer}.
The construction defines an analytical probability law only:
this reference process neither computes $b_k$ here nor performs the
replacement. The lemma is not needed to certify the precision-stopped
Algorithm~\ref{alg:sequential-newton-cg}.
whose recovery directly enforces the center bound.
\end{proof}

\subsection{Proofs of the four certification theorems}
\label{app:four-certification-proofs}
Each proof below establishes all three conclusions of its theorem in
one place. The tracking and risk arguments fix $D\in\mathcal G_n$ and
use the common notation and fixed-order setup of Appendix~\ref{app:tracking},
the tolerances in Appendix~\ref{app:hidden-proof}, and, for public
states, the original innovations in Appendix~\ref{app:three-step-proof}.
The shared risk proposition and finite-recovery lemma are in
Appendices~\ref{app:prd-proof} and~\ref{app:recovery-procedure}.
Certification uses the actual laws: its all-history recovery argument
extends to every finite dataset for which the global objective
assumptions hold. Data confidence and fast-stopping probabilities
are separate claims.
No unsafe-center conditioning, rejection of noise, or exact-fit
replacement is executed. Only the model releases are certified;
observable iteration counts, recovery flags, timing or aborts would
require an additional analysis.

\subsubsection{Pointwise partial certification}
\label{app:proof-hidden-one-step}
\begin{proof}[Proof of Theorem~\ref{thm:hidden-one-step}]
\textbf{Tracking and stopping.}\quad
Fix $D\in\mathcal G_n$. All estimates below hold uniformly over
$1\leq k\leq K$. The deterministic envelopes satisfy
$\tau_n^{(1)}=\widetilde O(n^{-1/2})$ and
$a_n=\widetilde O(n^{-1/2})$, while $b_n$ and
$\kappa_n=L/\mu\geq1$ have fixed polylogarithmic growth.
Consequently, for all sufficiently large $n$,
\[
 \tau_n^{(1)}\leq1,\qquad
 a_n+b_n\tau_n^{(1)}\leq\frac{1}{8\kappa_n},\qquad
 c_n\leq\frac{\tau_n^{(1)}}{4\kappa_n}.
\]
The last inequality follows directly from
\eqref{eq:two-step-center-tolerance}. The sufficiently-large-$n$
thresholds use only the fixed deterministic envelopes and are uniform
over $D\in\mathcal G_n$ and $k\leq K$.

Suppose the preceding accepted hidden center, which is the next input
$u$, satisfies $\|u-b_{k-1}\|\leq\tau_n^{(1)}$.
The deletion-shift bound \eqref{eq:oracle-newton-rewrite} gives
\[
 \|u-b_k\|\leq D_0+\tau_n^{(1)}<R_n,\qquad
 \|b_{k-1}-b_k\|\leq D_0<R_n.
\]
Thus both inputs lie in the local ball required by
Lemma~\ref{lem:accum_err}. Set $u_N=\mathcal N_k(u)$.
The Newton-map difference estimate established in the proof of
Lemma~\ref{lem:accum_err} gives
\[
\begin{aligned}
 \|u_N-\mathcal N_k(b_{k-1})\|
 &\leq a_n\|u-b_{k-1}\|+b_n\|u-b_{k-1}\|^2\\
 &\leq\frac{\tau_n^{(1)}}{8\kappa_n}.
\end{aligned}
\]
Combining this with the deletion remainder in
\eqref{eq:oracle-newton-rewrite} yields
\[
 \|u_N-b_k\|
 \leq c_n+\frac{\tau_n^{(1)}}{8\kappa_n}
 \leq\frac{3\tau_n^{(1)}}{8\kappa_n}.
\]

For the actual CG direction $d$, let
$r=g_k(u)-H_k(u)d$ and $e=H_k(u)^{-1}r$.
The true-residual rule \eqref{eq:true-cg-residual} gives
\[
 \|r\|\leq\frac{\mu\tau_n^{(1)}}{8},\qquad
 \|e\|\leq\frac{\tau_n^{(1)}}{8},\qquad
 u-d=u_N+e.
\]
When $g_k(u)=0$, the prescribed zero direction has $r=e=0$ and
obeys the same bounds. Before using any Hessian expansion, note that,
for every $s\in[0,1]$,
\[
 \|u_N+se-b_k\|
 \leq\frac{3\tau_n^{(1)}}{8\kappa_n}+\frac{\tau_n^{(1)}}{8}
 \leq\frac{\tau_n^{(1)}}{2}<R_n.
\]
Hence the exact Newton output, the actual CG output, and the entire
segment between them belong to the local ball. The segment from $b_k$
to $u_N$ also lies in that ball, so
\eqref{eq:local-spectrum-rewrite} implies
\[
 \|g_k(u_N)\|\leq L\|u_N-b_k\|
 \leq\frac{3\mu\tau_n^{(1)}}{8}.
\]

It remains to bound how the CG residual changes this gradient.
Using the existing quantity
$m_k(w)=\max\{\|X_kw\|_\infty,\|w\|_\infty\}$,
we have $m_k(w)\leq B_X\|w\|$ for every $w$.
Equation~\eqref{eq:rho-k-definition} bounds
$m_k(\mathcal N_k(b_{k-1})-b_{k-1})$ by $\rho_n$.
The decomposition
\[
\begin{aligned}
 u_N+se-u
 ={}&\mathcal N_k(b_{k-1})-b_{k-1}
     +\bigl(u_N-\mathcal N_k(b_{k-1})\bigr)\\
    &-(u-b_{k-1})+se
\end{aligned}
\]
therefore gives, uniformly over $s\in[0,1]$,
\[
 m_k(u_N+se-u)
 \leq\rho_n+B_X\left(
    \frac{\tau_n^{(1)}}{8\kappa_n}+\tau_n^{(1)}+\frac{\tau_n^{(1)}}{8}
 \right).
\]
Both arguments are in the local ball, so
\eqref{eq:local-directional-rewrite} and
$\|H_k(u)^{-1}\|_{\mathrm{op}}\leq\mu^{-1}$ imply
\[
\begin{aligned}
 &\sup_{s\in[0,1]}
 \bigl\|[H_k(u_N+se)-H_k(u)]H_k(u)^{-1}\bigr\|_{\mathrm{op}}\\
 &\qquad\leq\frac{A}{\mu}\left\{
    \rho_n+B_X\left(
      \frac{\tau_n^{(1)}}{8\kappa_n}+\tau_n^{(1)}+\frac{\tau_n^{(1)}}{8}
    \right)
 \right\}
 =\widetilde O(n^{-1/2})\leq1
\end{aligned}
\]
for all sufficiently large $n$. This bound is uniform over the
admissible inputs, since $A,B_X,\rho_n$, and $\tau_n^{(1)}$ are
deterministic envelopes. Taylor integration, with the matrix factors
in their displayed order, now gives the exact identity
\[
 g_k(u_N+e)-g_k(u_N)
 =\left[
    I_p+\int_0^1
    [H_k(u_N+se)-H_k(u)]H_k(u)^{-1}\,ds
  \right]r.
\]
Consequently the actual post-update gradient satisfies
\[
 \|g_k(u-d)\|
 \leq\frac{3\mu\tau_n^{(1)}}{8}+2\|r\|
 \leq\frac{5\mu\tau_n^{(1)}}{8}<\mu\tau_n^{(1)}.
\]
The first Newton--CG correction therefore passes the gradient gate,
and its center satisfies $\|u-d-b_k\|\leq\tau_n^{(1)}/2$ by the earlier
distance bound. Exact initialization starts the induction at $b_0$,
and this estimate maintains the required preceding-center bound
through all $k\leq K$. Thus $N_k=1$ at every deletion, recovery is
never invoked on $\mathcal G_n$, and the same conclusion holds for
every prespecified $T\leq K$. Publication noise does not enter this
hidden-state induction.

\par\medskip\noindent\textbf{Certification.}\quad
Lemma~\ref{lem:recovery-termination} gives a finite measurable center
with $\|C_k-b_k\|\leq\tau_n^{(1)}$ for every finite history, including
histories on which recovery is used. This uses global objective
regularity and the positive tolerance, without conditioning on tracking
success. The current Gaussian draw is independent of the center and
the past. Apply the single-release conclusion of
Theorem~\ref{thm:gaussian-certification} to each actual marginal.
With $\sigma_n=2\tau_n^{(1)}\sqrt{2\log(1/\delta)}/\epsilon$,
$v=(\tau_n^{(1)})^2/\sigma_n^2=\epsilon^2/(8\log(1/\delta))$.
The Gaussian threshold with $\delta_G=\delta$ is
\begin{equation}\label{eq:pointwise-gaussian-calibration}
 \epsilon/2+\epsilon^2/(16\log(1/\delta))\leq\epsilon,
\end{equation}
for the stated privacy parameters. Thus every actual released marginal
obeys both $(\epsilon,\delta)$ inequalities, proving pointwise partial
certification for the full actual laws.

\par\medskip\noindent\textbf{Population-risk degradation.}\quad
Fix the constant $d>1$ in the theorem and apply
Proposition~\ref{prop:prd-control} with
$\tau=\tau_n^{(1)}$, $\sigma=\sigma_n$,
$\rho_{\rm track}=0$, and $\rho_R=n^{-d}$. The deterministic hidden
tracking argument supplies its center-error hypothesis. On
$\mathcal G_n$, $B_*/\sqrt p\leq B/\sqrt p$ is polylogarithmic.
With $q=\sqrt{2\log(2T/\rho_R)}$ and
$W=\tau+\sigma(\sqrt p+q)$, the prefactor in
\eqref{eq:prd-explicit-bound} is also polylogarithmic for fixed privacy
parameters. Since $\tau,\sigma=\widetilde O(n^{-1/2})$ and $p\asymp n$,
the three terms $\tau/\sqrt p$, $\sigma q/\sqrt p$, and $W^2/p$
are each $\widetilde O(n^{-1})$. The proposition therefore gives
$\operatorname{PRD}_{1:T}=\widetilde O(n^{-1})$ with conditional
failure probability at most $n^{-d}$, as claimed. Publication
randomness remains in this realized risk; only the independent test
pair is averaged in $R$.
\end{proof}

\subsubsection{Pathwise partial certification}
\label{app:proof-hidden-two-step}
\begin{proof}[Proof of Theorem~\ref{thm:hidden-two-step-pathwise}]
\textbf{Tracking and stopping.}\quad
The explicit choice \eqref{eq:two-step-center-tolerance} satisfies,
eventually,
\begin{equation}\label{eq:two-step-smallness}
 \tau_n^{(2)}\leq w_n,\qquad 2w_n\leq1,\qquad
 a_n\tau_n^{(2)}+b_n(\tau_n^{(2)})^2+\tau_n^{(2)}/8\leq w_n.
\end{equation}
Indeed $w_n\geq n^{-1/2}$ and each numerator is a fixed
polylogarithm of the displayed order. If the prior error is at most
$\tau_n^{(2)}$, the first update has error $e_1\leq c_n+w_n\leq2w_n$.
If its gate accepts, strong convexity finishes the step. Otherwise,
\eqref{eq:quadratic-newton-rewrite} and the residual rule imply
\[
 e_2\leq b_n(2w_n)^2+\tau_n^{(2)}/8
       =2(L_H/\mu)w_n^2+\tau_n^{(2)}/8\leq\tau_n^{(2)}/4.
\]
This puts the second output in the ball before any gradient expansion.
The actual direction obeys
$\|d_2\|\leq e_1+e_2\leq2w_n+\tau_n^{(2)}/4\leq9w_n/4$.
Both endpoints and their segment are in the ball, so
\begin{equation}\label{eq:two-step-gradient-proof}
 \|g_k(u_2)\|\leq\mu\tau_n^{(2)}/8+(81/32)L_Hw_n^2
 \leq(1/8+81/512)\mu\tau_n^{(2)}<\mu\tau_n^{(2)}.
\end{equation}
This includes $L_H=0$. Thus the actual gate passes within two updates.
Exact initialization and induction give the full hidden path result.
Here $B_{\mathrm N}=J_\star=2$, and the second post-update gate is
checked before recovery can be invoked. Thus no recovery is reached
on $\mathcal G_n$.

\par\medskip\noindent\textbf{Certification.}\quad
Lemma~\ref{lem:recovery-termination} ensures a finite measurable center
within $\tau_n^{(2)}$ of $b_k$ at every finite history, whether or not
the fast tracking event holds. The current Gaussian noise is fresh and
independent of that center and the past. Apply
Theorem~\ref{thm:gaussian-certification} directly to the actual joint
release kernels. With
$\sigma_n=2\tau_n^{(2)}\sqrt{2T\log(1/\delta)}/\epsilon$, the joint
quantity is $v=T(\tau_n^{(2)})^2/\sigma_n^2=\epsilon^2/(8\log(1/\delta))$,
so, with $\delta_G=\delta$,
\[
 v/2+\sqrt{2v\log(1/\delta)}
 =\epsilon/2+\epsilon^2/(16\log(1/\delta))\leq\epsilon.
\]
This proves both $(\epsilon,\delta)$ inequalities for the complete
actual transcript, hence pathwise partial certification, without
conditioning its law on tracking success.

\par\medskip\noindent\textbf{Population-risk degradation.}\quad
For any fixed $d>0$, use Proposition~\ref{prop:prd-control} with
$\tau=\tau_n^{(2)}$, $\sigma=\sigma_n$,
$\rho_{\rm track}=0$ and $\rho_R=n^{-d}$. The hidden center bound
holds deterministically on each fixed $D\in\mathcal G_n$.
Set $q=\sqrt{2\log(2T/\rho_R)}$ and
$W=\tau+\sigma(\sqrt p+q)$. The good-data bound
$B_*/\sqrt p\leq B/\sqrt p$ makes the prefactor in
\eqref{eq:prd-explicit-bound} polylogarithmic. Substituting
$\tau=\widetilde O(n^{-1})$ and
$\sigma=\widetilde O(\sqrt T/n)$, its terms
$\tau/\sqrt p$, $\sigma q/\sqrt p$, and $W^2/p$ give
\[
 \operatorname{PRD}_{1:T}
 \leq\widetilde O(n^{-3/2}+\sqrt T/n^{3/2}+T/n^2)
 =\widetilde O(\sqrt T\,n^{-3/2})
\]
with conditional probability at least $1-n^{-d}$. The remaining square
terms are no larger, and the last comparison uses the prescribed
$1\leq T\leq K<n$. This is \eqref{eq:path-prd-rate}, including $T=1$
and linear $T$. The expectation in $R$ averages the test pair only;
the bound is for realized risk under publication noise.
\end{proof}

\subsubsection{Pointwise perfect certification}
\label{app:proof-public-two-step}
\begin{proof}[Proof of Theorem~\ref{thm:public-two-step-pointwise}]
\textbf{Tracking and stopping.}\quad
At its scale $\sigma_n=2\tau_n^{(2)}\sqrt{2\log(2/\delta)}/\epsilon$, set
$E_n=\tau_n^{(2)}+U_N$. For fixed $\rho$, or $\rho=n^{-d}/2$ with fixed
$d>0$, uniformly over $T\leq K$,
\[
 E_n=\widetilde O(n^{-1/2})=o(1),\qquad
 a_nE_n+b_nE_n^2+\tau_n^{(2)}/8=o(n^{-1/2})\leq w_n.
\]
On \eqref{eq:innovation-event}, assume the preceding actual centers
passed their gates. The next start differs from $b_{k-1}$ by at most
$E_n$ and lies within $D_0+E_n<R_n$ of $b_k$. The first output error
is at most $c_n+w_n\leq2w_n$. The same second-output bound and direct
gradient calculation \eqref{eq:two-step-gradient-proof} apply.
Explicitly, the eventual smallness conditions
\eqref{eq:two-step-smallness} and the same true-residual rule give,
if the first gate has not accepted,
\[
 e_2\leq b_n(2w_n)^2+\tau_n^{(2)}/8
 =2(L_H/\mu)w_n^2+\tau_n^{(2)}/8\leq\tau_n^{(2)}/4,
 \qquad \|d_2\|\leq e_1+e_2\leq9w_n/4.
\]
The second output, both endpoints and their segment are in the local
ball before applying the gradient bound. Hence
\[
 \|g_k(u_2)\|\leq\mu\tau_n^{(2)}/8+(81/32)L_Hw_n^2
 \leq(1/8+81/512)\mu\tau_n^{(2)}<\mu\tau_n^{(2)}.
\]
This calculation also includes $L_H=0$.
Earlier acceptance also implies error at most $\tau_n^{(2)}$.
The base case has exact center and $Z_0=0$, so induction covers the
actual process through $T$. Since $B_{\mathrm N}=J_\star=2$ and the
second post-update gate passes whenever it is reached, recovery is
not invoked on \eqref{eq:innovation-event}.
Its fast-tracking failure probability is at most $\rho$ on every fixed
good dataset; recovery still bounds final center error off this event.

\par\medskip\noindent\textbf{Certification.}\quad
For every finite history, Lemma~\ref{lem:recovery-termination} supplies
a finite measurable center satisfying
$\|C_k-b_k\|\leq\tau_n^{(2)}$. The fresh current Gaussian draw is
independent of this center and the past. Thus the single-release
bounded-center mixture conclusion of
Theorem~\ref{thm:gaussian-certification} applies to each actual public
marginal, including histories outside the fast-tracking event.
Use $\sigma_n=2\tau_n^{(2)}\sqrt{2\log(2/\delta)}/\epsilon$ and
put $L_2=\log(2/\delta)$. With $v=(\tau_n^{(2)})^2/\sigma_n^2
=\epsilon^2/(8L_2)$ and $\delta_G=\delta$, the threshold is
\[
 \frac{\epsilon^2}{16L_2}
 +\frac\epsilon2\sqrt{\frac{\log(1/\delta)}{L_2}}
 \leq\frac{\epsilon^2}{16\log(1/\delta)}+\frac\epsilon2
 \leq\epsilon.
\]
This proves bilateral pointwise perfect certification for the full
actual marginal laws, without conditioning away tracking failures.

\par\medskip\noindent\textbf{Population-risk degradation.}\quad
Fix $d>0$ and take $\rho=\rho_{\rm track}=n^{-d}/2$ in the tracking
argument above. Apply Proposition~\ref{prop:prd-control} with
$\tau=\tau_n^{(2)}$, $\sigma=\sigma_n$, and $\rho_R=n^{-d}/2$.
For $q=\sqrt{2\log(2T/\rho_R)}$ and
$W=\tau+\sigma(\sqrt p+q)$, the bound
$B_*/\sqrt p\leq B/\sqrt p$ on $\mathcal G_n$ and the fixed privacy
parameters make the prefactor in \eqref{eq:prd-explicit-bound}
polylogarithmic. Here $\tau,\sigma=\widetilde O(n^{-1})$.
Consequently $\tau/\sqrt p$ and $\sigma q/\sqrt p$ are
$\widetilde O(n^{-3/2})$, while $W^2/p=\widetilde O(n^{-2})$.
Thus $\operatorname{PRD}_{1:T}=\widetilde O(n^{-3/2})$ on the
intersection of the tracking and risk events. Their conditional
failure probabilities sum to at most $n^{-d}$; no independence of
these events is needed. This gives the claimed joint fast-tracking
and realized-risk guarantee, with certification already established
for the actual laws. The expectation in $R$ removes test-pair
randomness only, not publication randomness.
\end{proof}

\subsubsection{Pathwise perfect certification}
\label{app:proof-public-three-step}
\begin{proof}[Proof of Theorem~\ref{thm:perfect-three-step}]
\textbf{Tracking and stopping.}\quad
Use exactly \eqref{eq:public-tau-choice}. For the same confidence
choices, uniformly over $T\leq K$, $U_N=O(1)$ and
$U_D=O(\sqrt{\log(en)/n})$. The predetermined slack
$R_n-D_0-2=\log(en)$ contains $U_N$ eventually.
On \eqref{eq:innovation-event}, assume preceding actual gates passed.
For the next input $u$ and $\Delta=u-b_{k-1}$,
$\|\Delta\|\leq n^{-1}+U_N$ and
$m_k(\Delta)\leq B_X/n+U_D$. Define
\begin{align}
 E_1&=c_n+(A/\mu)(B_X/n+U_D)
           \{(n^{-1}+U_N)/2+D_0\}+1/(8n),\notag\\
 E_2&=b_nE_1^2+1/(8n),\qquad
 E_3=b_nE_2^2+1/(8n).
 \label{eq:public-three-errors}
\end{align}
The directional first bound and successive quadratic bounds give
errors at most $E_1,E_2,E_3$, unless an earlier gate already accepts.
Their orders are $\widetilde O(n^{-1/2})$, $\widetilde O(n^{-1})$,
and $1/(8n)+\widetilde O(n^{-2})$. Each output is within distance one
for large $n$, and the initial point is within $D_0+n^{-1}+U_N<R_n$.
Thus every used iterate and the required segments stay in the ball.
If a third update is taken, its direction has norm at most $E_2+E_3$.
Consequently
\begin{equation}\label{eq:public-third-gradient}
 \|g_k(u_3)\|\leq\mu/(8n)+(L_H/2)(E_2+E_3)^2\leq\mu/n
\end{equation}
eventually. This closes the induction from exact initialization and
proves fast-tracking failure at most $\rho$. Here $B_{\mathrm N}=J_\star=3$;
\eqref{eq:public-third-gradient} ensures that the last allowed update
passes the gate whenever it is reached, so no recovery is executed on
this event. The off-event accuracy guarantee is supplied
separately by Lemma~\ref{lem:recovery-termination}.

\par\medskip\noindent\textbf{Certification.}\quad
The positive tolerance $\tau_n=1/n$ and
Lemma~\ref{lem:recovery-termination} give a finite measurable center
with $\|C_k-b_k\|\leq1/n$ at every finite history. Fresh current
Gaussian noise is independent of the center and past. Therefore
Theorem~\ref{thm:gaussian-certification} applies directly to the actual
joint release kernels, including histories requiring recovery.
Under the exact calibration \eqref{eq:public-tau-choice},
$v=T\tau_n^2/\sigma_n^2=\epsilon^2/(8\log(1/\delta))$, and
\[
 v/2+\sqrt{2v\log(1/\delta)}
 =\epsilon/2+\epsilon^2/(16\log(1/\delta))\leq\epsilon.
\]
The choice $\delta_G=\delta$ consequently gives both inequalities for
the full actual transcript, proving pathwise perfect certification
without conditioning on fast tracking.

\par\medskip\noindent\textbf{Population-risk degradation.}\quad
For any fixed $d>0$, choose $\rho=\rho_{\rm track}=n^{-d}/2$ in the
original-innovation tracking argument and $\rho_R=n^{-d}/2$ in
Proposition~\ref{prop:prd-control}. Use the exact
$\tau=1/n$ and $\sigma=2\sqrt{2T\log(1/\delta)}/(\epsilon n)$.
Put $q=\sqrt{2\log(2T/\rho_R)}$ and
$W=\tau+\sigma(\sqrt p+q)$. On $\mathcal G_n$,
$B_*/\sqrt p\leq B/\sqrt p$ is polylogarithmic, and so is the
prefactor in \eqref{eq:prd-explicit-bound} for fixed privacy parameters.
The three principal contributions $\tau/\sqrt p$,
$\sigma q/\sqrt p$, and $W^2/p$, with the remaining square terms
no larger, yield
\[
 \operatorname{PRD}_{1:T}
 =\widetilde O\left(n^{-3/2}+\frac{\sqrt T}{n^{3/2}}+\frac{T}{n^2}\right).
\]
The tracking and risk events intersect with conditional probability
at least $1-\rho_{\rm track}-\rho_R=1-n^{-d}$, without any independence
requirement. This gives the stated three-term risk and fast-stopping
guarantees through the prespecified horizon, including $T=1$ and
linear $T$. Publication randomness remains in the realized risk;
only the independent test pair is averaged in $R$.
\end{proof}

\section{Realized population-risk degradation}
\label{app:prd-proof}
The expectation defining $R$ averages only an independent test pair;
publication randomness remains in $\operatorname{PRD}_{1:T}$.

\begin{lemma}[Population derivatives]
\label{lem:population-risk-derivatives}
Under the stated test-population assumptions, $R$ is $C^2$ and, for a
constant $C_R$ depending only on fixed growth, covariance and response
moment constants,
\begin{align}
 \|\nabla R(\beta)\|&\leq
 \frac{C_R}{\sqrt p}\{1+(\|\beta\|/\sqrt p)^s\},
 \label{eq:population-gradient-bound}\\
 0\preceq\nabla^2R(\beta)&\preceq
 \frac{C_R}{p}\{1+(\|\beta\|/\sqrt p)^s\}I_p.
 \label{eq:population-hessian-bound}
\end{align}
No independence within the test pair is assumed.
\end{lemma}
\begin{proof}
The second-derivative identity in the proof of
Lemma~\ref{lem:lipschitz_hessian} implies globally
$0\leq\ddot\ell(y,t)\leq C_2(1+|y|^s+|t|^s)$.
For any unit $v$, Gaussian scalar moments give
\[
 \EE|x_0^\top\beta|^{2s}\leq C\|\beta\|^{2s}/p^s,\quad
 \EE|x_0^\top v|^2\leq C/p,\quad
 \EE|x_0^\top v|^4\leq C/p^2.
\]
The response moment bounds the squared first and second loss
derivatives in expectation by $C\{1+(\|\beta\|/\sqrt p)^{2s}\}$.
On a compact parameter ball the first and second differentiation
envelopes are constant multiples of
$(1+|y_0|^s+\|x_0\|^s)\|x_0\|$ and
$(1+|y_0|^s+\|x_0\|^s)\|x_0\|^2$.
Cauchy--Schwarz and Gaussian moments make both integrable, so
dominated convergence justifies differentiation and continuity.
Then Cauchy--Schwarz applied to
\[
 v^\top\nabla R(\beta)=\EE[\dot\ell(y_0,x_0^\top\beta)x_0^\top v],
 \quad
 v^\top\nabla^2R(\beta)v
 =\EE[\ddot\ell(y_0,x_0^\top\beta)(x_0^\top v)^2]
\]
gives the factors $p^{-1/2}$ and $p^{-1}$ respectively. Convexity
gives positivity, and taking the supremum over unit $v$ proves both
bounds.
\end{proof}

\begin{proposition}[Full-path realized risk]
\label{prop:prd-control}
Fix data and let $B_*=\max_{k\leq T}\|b_k\|$.
Suppose the actual center path has error at most $\tau$ except on an
event of conditional probability at most $\rho_{\rm track}$.
For $\rho_R\in(0,1)$ put $q=\sqrt{2\log(2T/\rho_R)}$ and
$W=\tau+\sigma(\sqrt p+q)$.
With conditional probability at least
$1-\rho_{\rm track}-\rho_R$,
\begin{equation}\label{eq:prd-explicit-bound}
 \operatorname{PRD}_{1:T}
 \leq C_R\{1+(B_*/\sqrt p)^s+(W/\sqrt p)^s\}
       \left\{\frac\tau{\sqrt p}+\frac{\sigma q}{\sqrt p}
                       +\frac{W^2}{p}\right\},
\end{equation}
after increasing $C_R$ only by a factor depending on $s$.
Hidden tracking permits $\rho_{\rm track}=0$.
\end{proposition}
\begin{proof}
Conditional on data, each $\nabla R(b_k)$ is deterministic. Under the
original Gaussian innovation law, the norm tail and one-sided scalar
tail, each unioned over $T$ rounds, give probability at least
$1-\rho_R$ to
\begin{align}
 \|Z_k\|&\leq\sigma(\sqrt p+q),
 \label{eq:prd-noise-norm}\\
 \langle\nabla R(b_k),Z_k\rangle
 &\leq\sigma\|\nabla R(b_k)\|q\quad\text{for all }k\leq T.
 \label{eq:prd-noise-linear}
\end{align}
The second bound is automatic for zero gradient. These events are
established before intersecting tracking, and independence between
the two events is not required. On their intersection write the
actual release perturbation as $e_k+Z_k$, with $\|e_k\|\leq\tau$.
Taylor's formula bounds its linear term by
$(\tau+\sigma q)\|\nabla R(b_k)\|$.
Every point on the Taylor segment has norm at most $B_*+W$.
Lemma~\ref{lem:population-risk-derivatives} therefore bounds the
quadratic remainder by the right-side prefactor times $W^2/p$.
This nonnegative bound applies to the positive part and to the maximum
over $k$. Intersecting events costs at most
$\rho_{\rm track}+\rho_R$, proving the proposition.
\end{proof}

On $\mathcal G_n$, $B_*/\sqrt p\leq B/\sqrt p$ is polylogarithmic.
For hidden mechanisms choose $\rho_R=n^{-d}$; for public mechanisms
choose $\rho_{\rm track}=\rho_R=n^{-d}/2$ in the actual tracking
and risk proofs. The first factor in \eqref{eq:prd-explicit-bound}
is then polylogarithmic for fixed privacy parameters. The substitutions
in the complete proofs in Appendix~\ref{app:four-certification-proofs}
are summarized below:
\begin{center}
\begin{tabular}{lll}
Setting & $(\tau,\sigma)$, up to logarithms & $\operatorname{PRD}_{1:T}$\\ \hline
Hidden pointwise & $(n^{-1/2},n^{-1/2})$ & $\widetilde O(n^{-1})$\\
Hidden pathwise & $(n^{-1},\sqrt T/n)$ & \eqref{eq:path-prd-rate}\\
Public pointwise & $(n^{-1},n^{-1})$ & $\widetilde O(n^{-3/2})$\\
Public pathwise & $(n^{-1},\sqrt T/n)$ & \eqref{eq:path-prd-rate}
\end{tabular}
\end{center}
For example the three principal pathwise terms are
$\tau/\sqrt p$, $\sigma q/\sqrt p$ and $\sigma^2$,
giving $\widetilde O(n^{-3/2}+\sqrt T/n^{3/2}+T/n^2)$; the remaining
square terms are no larger. These are the PRD substitutions used in
the four complete proofs, including $T=1$ and linear $T$.
Data-and-noise success is at least $1-p_n-\rho_{\rm track}-\rho_R$.
These are accuracy events, whereas $\delta$ is the slack in a
distributional comparison. In particular a public parameter noise
vector can have nonvanishing Euclidean norm while this realized
population-risk bound tends to zero.

\section{Conjugate-gradient residuals and arithmetic work}
\label{app:cg-proof}

Recall we use \textproc{ConjugateGrad} (Algorithm~\ref{alg:cg}) to
approximately solve the Newton direction $d=H^{-1}g$. For each
Newton--CG update, set $\eta=\mu\tau_n/8$, as in Section~2.2, and
return the first iterate satisfying $\|g-Hd\|_2\leq\eta$.
The initial iterate $d_0=0$ is included in this test. Thus, if
$\|g\|_2\leq\eta$, the solver returns $d=0$ without a CG iteration,
and the subsequent outer gradient gate passes immediately. This is
an inner-solver stopping convention; the outer update and its gate
are still counted as prescribed by Algorithm~\ref{alg:sequential-newton-cg}.

\begin{algorithm}[t]
\caption{\textproc{ConjugateGrad}: solving $Hd=g$}
\label{alg:cg}
\KwIn{Symmetric positive-definite matrix $H\in\RR^{p\times p}$; vector $g\in\RR^p$; tolerance $\eta>0$}
\KwOut{Direction $d$ satisfying $\|g-Hd\|_2\leq\eta$}
Set $d_0=0$, $r_0=g$, $q_0=r_0$, and $m=0$\;
\While{$\|g-Hd_m\|_2>\eta$}{
 Compute $v_m=Hq_m$ and
 $\alpha_m=(r_m^\top r_m)/(q_m^\top v_m)$\;
 Set $d_{m+1}=d_m+\alpha_m q_m$ and
 $r_{m+1}=r_m-\alpha_m v_m$\;
 Set $q_{m+1}=r_{m+1}
       +\dfrac{r_{m+1}^\top r_{m+1}}{r_m^\top r_m}q_m$\;
 Set $m=m+1$\;
}
\Return{$d_m$}\;
\end{algorithm}

We use an exact real-arithmetic oracle model. Arithmetic
operations and comparisons, scalar evaluations of the loss
and each penalty term and their first two derivatives, and
draws of a standard Gaussian scalar each have unit cost,
uniformly over finite inputs. Initial training is excluded.
These bounds concern arithmetic work, not bit complexity
or measured runtime.

\begin{proposition}[True residual and actual work]
\label{prop:cg-work}
For an SPD Newton Hessian of condition number $\kappa>1$, zero-start
CG in exact arithmetic obeys
\begin{equation}\label{eq:cg-residual-bound}
 \|r_m\|\leq2\sqrt\kappa
 \left(\frac{\sqrt\kappa-1}{\sqrt\kappa+1}\right)^m\|g\|.
\end{equation}
Let $m_{\mathrm{stop}}$ be the first iteration index $m\geq0$
satisfying the absolute true-residual test
\begin{equation}\label{eq:true-cg-residual}
 \|g_{\backslash k}(u)-H_{\backslash k}(u)d_m\|_2
 \leq\mu\tau_n/8.
\end{equation}
In particular, $m_{\mathrm{stop}}=0$ whenever
$\|g\|_2\leq\mu\tau_n/8$, including $g=0$. For $\kappa>1$, we have
\begin{equation}\label{eq:cg-iteration-bound}
 m_{\mathrm{stop}}\leq\min\left\{p,\left\lceil\frac{\sqrt\kappa}{2}
       \log\left(2\sqrt\kappa\max\{1,8\|g\|/(\mu\tau_n)\}\right)
                        \right\rceil\right\}.
\end{equation}
For $\kappa=1$, one iteration suffices whenever the initial residual
fails the test; otherwise $m_{\mathrm{stop}}=0$.
On the four tracking events, all actual solves require
$O(1+\sqrt\kappa\log n)=\polylog(n)$ iterations. Dense post-training
work is $\widetilde O(np)$ per deletion and $\widetilde O(Tnp)$
overall, with $O(np)$ data storage and $O(n+p)$ additional workspace.
\end{proposition}
\begin{proof}
Let $d^*=H^{-1}g$ and $\|v\|_H^2=v^\top Hv$. Exact CG's
$H$-conjugate directions and orthogonal residuals make it minimize
the quadratic over its Krylov space. Its error energy is therefore
no greater than that obtained by any degree-$m$ polynomial $P(H)d^*$
with $P(0)=1$. Let $\lambda_-$ and $\lambda_+$ be the extreme
eigenvalues. For $\kappa>1$ use the Chebyshev polynomial
\[
 P(x)=\frac{T_m((\lambda_++\lambda_--2x)/(\lambda_+-\lambda_-))}
 {T_m((\lambda_++\lambda_-)/(\lambda_+-\lambda_-))}.
\]
The numerator has absolute value at most one on the spectral interval.
For $u>1$,
$T_m(u)=\{(u+\sqrt{u^2-1})^m+(u-\sqrt{u^2-1})^m\}/2$.
At the denominator argument,
$u+\sqrt{u^2-1}=(\sqrt\kappa+1)/(\sqrt\kappa-1)$.
Thus the error energy norm is at most twice the ratio in
\eqref{eq:cg-residual-bound} to the power $m$, times $\|d^*\|_H$.
Using $\|r_m\|\leq\sqrt{\lambda_+}\|d_m-d^*\|_H$ and
$\|d^*\|_H\leq\|g\|/\sqrt{\lambda_-}$ proves the residual bound.
If $\|g\|\leq\mu\tau_n/8$, the initial iterate already passes the
absolute residual test, so $m_{\mathrm{stop}}=0$ and
\eqref{eq:cg-iteration-bound} holds. Otherwise the initial-to-target
ratio is $8\|g\|/(\mu\tau_n)>1$. Since
$\log((\sqrt\kappa+1)/(\sqrt\kappa-1))\geq2/\sqrt\kappa$,
the ceiling in \eqref{eq:cg-iteration-bound} makes the residual bound
at most $\mu\tau_n/8$. The maximum with $1$ merely gives a single
nonnegative upper bound covering the zero-iteration case; it imposes
no relative-residual requirement. Exact CG also terminates in at most
$p$ iterations. For a scalar multiple of the identity ($\kappa=1$),
one iteration gives the exact solution if the initial test failed.

The tracking proofs place every called Hessian at a parameter in
$B(b_k,R_n)$. Hence $\kappa\leq L/\mu=\polylog(n)$ and, by
integrating the Hessian toward $b_k$, $\|g\|\leq LR_n=\polylog(n)$.
Every selected $\tau_n$ is at least $1/n$, and $\mu>0$ is fixed,
so the initial-to-target ratio is polynomially bounded in $n$. The local-ball condition applies to the
parameter where the Hessian is frozen, not to the Krylov vectors.

For dense data and constant-cost scalar derivatives,
\[
 H_k(u)v=X_k^\top\operatorname{diag}
   (\ddot\ell(y_i,x_i^\top u))X_kv
       +\lambda\operatorname{diag}(r_j''(u_j))v
\]
costs $O((n-k)p+p)$ operations; a gradient has the same cost.
True-residual recomputation uses at most one further Hessian product
per iteration. At most $J_\star\leq3$ corrections on the tracking event, gradient gates and $O(p)$ noise
generation give the claimed total orders. Retained-data storage uses at most $O(np)$ space; deleted rows are
removed and are unavailable to later updates. Predictors, weights
and a constant number of Krylov vectors use $O(n+p)$ workspace.
Initial training costs are additional.
\end{proof}
For hidden updates these work bounds are deterministic on
$\mathcal G_n$. For public updates they hold on the original-noise
tracking event, with any stated fixed polynomial conditional failure
budget for sufficiently large $n$. Without that event, SPD exact CG still takes at most $p$ iterations
per solve. The finite Newton prefix therefore costs $O(B_{\mathrm N}Tnp^2)$,
where $B_{\mathrm N}=J_\star\leq3$, before any recovery work. The full cost,
including recovery, is treated in Proposition~\ref{prop:guarded-work}.
Neither arithmetic bound is a finite-precision or measured-runtime claim.

\section{Capped references and stopping-count sharpness}
\label{app:prescribed-update-sharpness}
This appendix analyzes capped reference versions of
Algorithm~\ref{alg:sequential-newton-cg}. They use the same state rules
and CG residual rule~\eqref{eq:true-cg-residual}, but perform at most
$J$ Newton steps per deletion, stopping earlier if a post-update
gradient test passes. They publish the final capped center with the
prescribed Gaussian noise even if the last test fails, and perform no
recovery. Their centers and laws are local reference notation, distinct
from the guarded algorithm's post-recovery centers. The final subsection
transfers the relevant first failed prefixes to the precision-stopped
algorithms.

\begin{theorem}[Obstructions for capped reference processes]
\label{thm:prescribed-update-sharpness}
Fix $0<\epsilon\leq1$, $0<\delta<1/2$, $p/n\to\gamma\in(0,1)$,
and fixed positive $t_-,c'$ with $c'<1$. Let the prespecified horizon
satisfy $t_-n\leq T\leq\lfloor c'n\rfloor$ eventually, with exact,
noiseless initialization and a deletion order independent of the data
and mechanism randomness. For each item below, there exist a fixed
model family satisfying Assumptions~\ref{ass:design}--\ref{ass:Penalty}
and data-measurable events $E_n\subseteq\mathcal G_n$, with
$\mathbb P_D(E_n)\to1$, such that for all sufficiently large $n$ and
every fixed $D\in E_n$, the indicated actual output law fails the
bilateral $(\epsilon,\delta)$ comparison with its independent
chronological retraining reference. Models may differ between items;
all model parameters are fixed before observing data or mechanism
noise. Thresholds may depend on the fixed model, privacy parameters,
dimension and horizon bounds, data-confidence parameter, and fixed
deterministic envelope choices.
\begin{enumerate}[label=(\roman*),leftmargin=*,itemsep=2pt]
\item \emph{Pathwise partial certification.}
Use the hidden capped reference with $j=1$ and the calibration of
Theorem~\ref{thm:hidden-two-step-pathwise}, retaining the complete
$\tau_n^{(2)}$ in \eqref{eq:two-step-center-tolerance} and that theorem's
Gaussian noise scale. The actual hidden-state release transcript
fails pathwise partial certification.
\item \emph{Pointwise perfect certification.}
Use the public capped reference with $j=1$ and the calibration of
Theorem~\ref{thm:public-two-step-pointwise}, retaining its
$\tau_n^{(2)}$ and Gaussian noise scale. The first actual release marginal
already fails pointwise perfect certification.
\item \emph{Pathwise perfect certification.}
For the public capped reference, retain
\eqref{eq:public-tau-choice} from Theorem~\ref{thm:public-three-step}.
The actual public transcript fails pathwise perfect certification
for each of the caps $j=2$ and $j=1$.
\end{enumerate}
In every item, the true-residual CG rule, post-update early stopping,
and release of the final capped center are unchanged; only its cap changes.
No recovery is performed in these reference processes.
\end{theorem}

Write $b_k=\beta^*_{\backslash k}$, $F_k=F_{\backslash k}$,
$g_k=\nabla F_k$, $H_k=\nabla^2F_k$, and
$\mathcal N_k(u)=u-H_k(u)^{-1}g_k(u)$.
Vector norms without subscripts are Euclidean and matrix norms are
operator norms. The notation $u^{\circ2}$ denotes coordinatewise
squaring, $\mathbf1$ the all-ones vector, and
$\operatorname{sech}(t)=1/\cosh(t)$.
Fix $0<\gamma_-<\gamma<\gamma_+<1$ so that
$\gamma_-\leq p/n\leq\gamma_+$ eventually. In particular $p\leq n$
eventually. Fix the prescribed deletion order and relabel it as
$1,\ldots,T$ by exchangeability. For an independently randomized
order, first condition on its realization, as in
Appendix~\ref{app:data-event}; no simultaneous statement over all
permutations is required.

For the two models below the response-tail probabilities can be zero,
so the existing class $\mathcal G_n$ obeys
$\mathbb P_D(\mathcal G_n^c)\leq n^{-d}$ for fixed $d>0$.
We intersect our data events with this class after proving their
probability bounds. Data failure probabilities remain separate from
the fixed-data certification slack $\delta$.
For parts (i) and (ii), the complete prescribed tolerance and variances are
\begin{equation}\label{eq:sharp-two-calibrations}
\begin{aligned}
\tau_n^{(2)}&=(1+16L_H/\mu)(c_n+n^{-1/2})^2,
& n^{-1}\leq\tau_n^{(2)}&\leq C_\tau\log(en)^{K_\tau}/n,\\
\sigma_H^2&=8T(\tau_n^{(2)})^2\log(1/\delta)/\epsilon^2,
&\sigma_P^2&=8(\tau_n^{(2)})^2\log(2/\delta)/\epsilon^2.
\end{aligned}
\end{equation}
Here $C_\tau,K_\tau$ are finite constants for the fixed deterministic
envelopes in \eqref{eq:two-step-center-tolerance}; they are not fitted
to a realized path. Part (iii) uses exactly
\begin{equation}\label{eq:sharp-public-calibration}
\tau=1/n,\qquad \sigma^2=\kappa_\sigma T/n^2,
\qquad \kappa_\sigma=8\log(1/\delta)/\epsilon^2.
\end{equation}
All sufficiently-large-$n$ thresholds below are taken after fixing
the model, privacy parameters, dimension and horizon bounds,
$d$, and the deterministic envelopes.

\subsection{Actual updates and distributional comparison tools}
\label{app:sharp-tools}
For an actual CG direction $d$, set $r=g_k(u)-H_k(u)d$.
The true-residual rule \eqref{eq:true-cg-residual} and strong convexity give
\begin{equation}\label{eq:sharp-actual-cg}
C=u-d=\mathcal N_k(u)+H_k(u)^{-1}r,
\qquad \|C-\mathcal N_k(u)\|\leq\tau_n/8.
\end{equation}
This bound applies to the actual returned residual, regardless of its
direction. Exact-arithmetic CG terminates in at most $p$ iterations
by Proposition~\ref{prop:cg-work}; an initial residual satisfying
$\|g_k(u)\|\leq\mu\tau_n/8$ returns $d=0$ without a CG iteration.
The bound \eqref{eq:sharp-actual-cg} also covers this return, since
then $r=g_k(u)$. Its finite operations and measurable comparison branches,
followed by a finite outer cap, make the actual centers finite
measurable functions of the preceding releases. Also,
Lemma~\ref{lem:uniform_strong_convexity} gives
\begin{equation}\label{eq:sharp-gate-lower}
\|g_k(u)\|\geq\mu\|u-b_k\|.
\end{equation}
Thus a center error exceeding $\tau_n$ forces the actual gate to fail.
Distributional failure will be established separately below.

For two laws define
$\mathrm D_\epsilon(P,Q)=\sup_A\{P(A)-e^\epsilon Q(A)\}$.
If the laws have positive densities and $L=\log(dP/dQ)$, pointwise
maximization of the signed density gives
\begin{equation}\label{eq:sharp-hockey-stick}
\mathrm D_\epsilon(P,Q)=\mathbb E_P[(1-e^{\epsilon-L})_+],
\end{equation}
with maximizing event $\{L>\epsilon\}$.

\begin{lemma}[Deterministic Gaussian shifts]
\label{lem:sharp-gaussian-shift}
For $P=N(m,\sigma^2I)$ and $Q=N(b,\sigma^2I)$, put
$s=\|m-b\|/\sigma$. Both directed divergences equal
\begin{equation}\label{eq:sharp-gaussian-shift}
\Delta_\epsilon(s)=
\Phi(s/2-\epsilon/s)-e^\epsilon\Phi(-s/2-\epsilon/s)
\quad(s>0),\qquad \Delta_\epsilon(0)=0.
\end{equation}
The function is continuous, strictly increasing, and tends to one as
$s\to\infty$.
\end{lemma}
\begin{proof}
Completing the square gives
\[
L(y)=\frac{(m-b)^\top(y-(m+b)/2)}{\sigma^2}.
\]
Under $P$ and $Q$ this is respectively
$N(s^2/2,s^2)$ and $N(-s^2/2,s^2)$.
Integrating the positive signed density over $\{L>\epsilon\}$ proves
\eqref{eq:sharp-gaussian-shift}; exchanging the centers gives the
same reverse divergence. If $\phi$ is the standard normal density,
$e^\epsilon\phi(-s/2-\epsilon/s)=\phi(s/2-\epsilon/s)$, and direct
differentiation gives
$\Delta_\epsilon'(s)=\phi(s/2-\epsilon/s)>0$.
The limits at zero and infinity follow from the Gaussian tails.
\end{proof}
Given data, hidden centers are deterministic and may be concatenated
into a $pT$-dimensional Gaussian shift. The first public center is
also deterministic because initialization is exact and noiseless.
Later public centers depend on history and require the next lemma.

\begin{lemma}[An adapted transcript test]
\label{lem:sharp-adapted-test}
Let $P$ be the actual law of
$Y_k=f_k(Y_{<k})+Z_k$, where $f_k$ are finite measurable functions
and $Z_k$ are independent $N(0,\sigma^2I)$ innovations. For fixed
finite $b_k$, let $Q=\bigotimes_{k=1}^T N(b_k,\sigma^2I)$.
Under $P$, define
\[
a_k=f_k(Y_{<k})-b_k,\qquad
M_t=\sum_{k\leq t}\frac{a_k^\top Z_k}{\sigma^2},\qquad
V_t=\sum_{k\leq t}\frac{\|a_k\|^2}{\sigma^2}.
\]
Then $L_T=\log(dP/dQ)(Y_{1:T})=M_T+V_T/2$, and for every $v>0$,
\begin{equation}\label{eq:sharp-adapted-test}
\mathrm D_\epsilon(P,Q)\geq
(1-e^{\epsilon-v/4})_+
\bigl[P(V_T\geq v)-e^{-v/32}\bigr]_+.
\end{equation}
In particular, $V_T\to\infty$ in $P$-probability implies
$\mathrm D_\epsilon(P,Q)\to1$.
\end{lemma}
\begin{proof}
Compare the positive conditional densities at the same numerical
history. Chronological factorization and square completion yield
$L_T=M_T+V_T/2$ exactly. With
$\mathcal F_{k-1}=\sigma(Y_1,\ldots,Y_{k-1})$, the coefficient $a_k$
is measurable before the fresh $Z_k$ is drawn. The single-step factor
\[
R_k=\exp\!\left\{-\frac{a_k^\top Z_k}{4\sigma^2}
                  -\frac{\|a_k\|^2}{32\sigma^2}\right\}
\]
satisfies $\mathbb E_P[R_k\mid\mathcal F_{k-1}]=1$ by conditional
Gaussian integration. Consequently the cumulative process
$\mathcal M_k=\exp(-M_k/4-V_k/32)=\mathcal M_{k-1}R_k$ satisfies
\begin{equation}\label{eq:sharp-martingale}
\mathbb E_P[\mathcal M_k\mid\mathcal F_{k-1}]=\mathcal M_{k-1},
\qquad \mathcal M_0=1,\qquad \mathbb E_P\mathcal M_k=1.
\end{equation}
The conditional expectation of the single-step factor is one;
that of the cumulative process is its previous value. Nonnegativity
and finite-step conditional integration require no unconditional
moment bound on the center maps.
On $\{M_T\leq-V_T/4,V_T\geq v\}$, one has
$\mathcal M_T\geq e^{v/32}$, so Markov's inequality bounds this
event's probability by $e^{-v/32}$. On the remainder of
$\{V_T\geq v\}$, $L_T\geq v/4$.
Restricting \eqref{eq:sharp-hockey-stick} to that event proves
\eqref{eq:sharp-adapted-test}. For the last assertion, first hold
$v$ fixed and take a limit inferior, then let $v\to\infty$.
The argument never conditions on the whole future coefficient
sequence to assert that $M_T$ is Gaussian.
\end{proof}

\subsection{A deletion-driven model and its data geometry}
\label{app:sharp-hidden-model}
For parts (i) and (ii), take independent observation pairs with
$x_i\sim N(0,I_p/p)$ and independent fair Rademacher responses $y_i$.
Use $\lambda=1$, squared loss, and the separable penalty
\begin{equation}\label{eq:sharp-hidden-model}
F_k(\beta)=\frac12\sum_{i>k}(y_i-x_i^\top\beta)^2
             +\sum_{j=1}^p q(\beta_j),\qquad
q'(t)=2t+\log\cosh t,\quad q(0)=0.
\end{equation}
Here $q'(0)=0$, $1\leq q''(t)=2+\tanh t\leq3$, and
$q'''(t)=\operatorname{sech}^2t\in(0,1]$.
Thus $q(t)\geq t^2/2\geq0$, $\nu=\mu=1$, the penalty Hessian has
global operator bound three, and
\[
\|\nabla^2r(u)-\nabla^2r(v)\|_{\mathrm F}^2
\leq\sum_j|u_j-v_j|^2.
\]
Squared loss is nonnegative, convex and $C^3$, has the required
quadratic growth, and has zero third derivative. Bounded responses
satisfy every required moment and tail condition; the independent
test pair has the same distribution. The design covariance is $I_p$.
These observations verify Assumptions~\ref{ass:design}--\ref{ass:Penalty}
for this fixed model. Comparing the objective at zero gives
$\|b_k\|^2\leq n-k\leq n$.

Define the deterministic quantities
\begin{equation}\label{eq:sharp-hidden-constants}
\begin{split}
M^2&=16\bigl[(1+1/\gamma_-)\log9+1\bigr],\qquad H=M^2+3,
\qquad B_{\rm fit}=\sqrt{2/\gamma_-},\\
h_n&=\sqrt{2(d+6)\log(en)/p},\qquad
u_n=\sqrt{2(d+6)\log(en)/\gamma_-},\qquad S_n=1+u_n.
\end{split}
\end{equation}
Put $S_k=\{j:|b_{k,j}|\leq B_{\rm fit}\}$ and
$w_k=H_k(b_k)^{-1}x_k$. The fit bound implies $|S_k|\geq p/2$.
Let $E_n$ be the intersection of $\mathcal G_n$ with
\begin{equation}\label{eq:sharp-hidden-data-event}
\begin{gathered}
\|X\|\leq M,\qquad
\|w_k\|_\infty\leq h_n,\qquad |x_k^\top b_k|\leq u_n,\\
\|(w_k)_{S_k}\|\geq(2H)^{-1}\qquad(1\leq k\leq T).
\end{gathered}
\end{equation}

\begin{lemma}[Data geometry for the deletion model]
\label{lem:sharp-hidden-geometry}
The event $E_n$ is data-measurable and satisfies, eventually,
\begin{equation}\label{eq:sharp-hidden-data-probability}
\mathbb P_D(E_n^c)\leq n^{-d}+2e^{-2p}
                    +4n^{-(d+4)}+ne^{-p/32}=o(1).
\end{equation}
\end{lemma}
\begin{proof}
A maximal $1/4$-separated subset of a unit sphere has at most
$9^r$ points in dimension $r$, by comparing disjoint radius-$1/8$
balls with a radius-$9/8$ ball, and is a $1/4$-net.
Approximating the two maximizing unit vectors gives
$\|X\|\leq2\max_{\rm nets}|v^\top Xu|$.
Every fixed bilinear form is Gaussian with variance $1/p$, so
\[
\mathbb P_D(\|X\|>M)
\leq2\exp\{(n+p)\log9-pM^2/8\}\leq2e^{-2p}.
\]
For each $k$, condition only on the retained pairs with indices
greater than $k$. Then $b_k,H_k(b_k),S_k$ are fixed and $x_k$ is
fresh Gaussian. Since $H_k(b_k)\succeq I$, each coordinate of $w_k$
has variance at most $1/p$, while $x_k^\top b_k$ has variance at most
$n/p\leq1/\gamma_-$. Scalar Gaussian tails and a union over at most
$np+n$ tests give total failure at most $4n^{-(d+4)}$.

Within this retained-data conditioning, restrict to the
retained-measurable event $\|X_k\|\leq M$.
Then $I\preceq H_k(b_k)\preceq HI$ and the covariance of
$(w_k)_{S_k}$ is at least $I/(pH^2)$. Diagonalization and the same
standard normal coordinates show that its squared norm dominates
$\chi^2_{|S_k|}/(pH^2)$. Since
\[
\mathbb P(\chi^2_r<r/2)
\leq e^{r/4}2^{-r/2}\leq e^{-r/16},
\]
the restricted norm fails with probability at most $e^{-p/32}$.
Integrate on retained spectral success, take a union over $k$, and
only then intersect with full spectral success, which implies all
retained spectral events. Finally intersect with $\mathcal G_n$ at
cost at most $n^{-d}$. This proves
\eqref{eq:sharp-hidden-data-probability} without conditioning a
deleted row on full-design spectral success. The events use only
measurable exact fits and data, not approximate iterates or noise.
\end{proof}

\subsection{Deletion error and the actual hidden centers}
\label{app:sharp-hidden-recursion}
\begin{lemma}[Residual-dependent error and inherited perturbations]
\label{lem:sharp-newton-error}
There are a fixed $c_*>0$ and deterministic
$A_n=O(\log(en)^2/\sqrt n)$,
$\rho_n=O(\log(en)^{3/2}/\sqrt n)$ such that, for each fixed
$D\in E_n$ and every $k\leq T$, writing $e_k=y_k-x_k^\top b_k$,
\begin{equation}\label{eq:sharp-newton-bounds}
\frac{c_*e_k^2}{\sqrt n}
\leq\|\mathcal N_k(b_{k-1})-b_k\|\leq A_n.
\end{equation}
For every vector $a$,
\begin{equation}\label{eq:sharp-map-bound}
\|\mathcal N_k(b_{k-1}+a)-\mathcal N_k(b_{k-1})\|
\leq\rho_n\|a\|+\tfrac12\|a\|^2.
\end{equation}
\end{lemma}
\begin{proof}
Fix $k$ and abbreviate $z=b_k$, $v=b_{k-1}$, $\Delta=v-z$,
$e=y_k-x_k^\top z$, $s=y_k-x_k^\top v$,
$H_0=H_k(z)$, $w=H_0^{-1}x_k$, and
$\bar H=\int_0^1H_k(z+t\Delta)\,dt$.
The optimality conditions give
\begin{equation}\label{eq:sharp-deletion-identity}
g_k(v)=sx_k,\qquad \bar H\Delta=sx_k,\qquad
e=s(1+x_k^\top\bar H^{-1}x_k).
\end{equation}
Hence $|e|/(1+M^2)\leq|s|\leq|e|\leq S_n$ and
$\|\Delta\|\leq M|s|$. The Hessian varies only through its diagonal
penalty, and $|(H_0-\bar H)_{jj}|\leq|\Delta_j|/2$.
The ordered resolvent identity therefore yields
\begin{equation}\label{eq:sharp-resolvent-shift}
\Delta-sw=s\bar H^{-1}(H_0-\bar H)w,\qquad
\|\Delta-sw\|\leq(M/2)h_ns^2.
\end{equation}
In particular the residual-dependent factor $s^2$ is retained.
Since $h_nS_n\to0$, uniformly over all such residuals and indices,
\begin{equation}\label{eq:sharp-restricted-shift}
\|\Delta_{S_k}\|\geq\frac{|s|}{4H},\qquad
\|\Delta\|_\infty\leq D_n:=S_nh_n+(M/2)S_n^2h_n=o(1)
\end{equation}
for all sufficiently large $n$.

Taylor integration gives the exact remainder
\begin{equation}\label{eq:sharp-newton-remainder}
\mathcal N_k(v)-z=H_k(v)^{-1}a^{\rm rem},\qquad
a^{\rm rem}_j=\Delta_j^2\int_0^1tq'''(z_j+t\Delta_j)\,dt.
\end{equation}
Every coordinate of $a^{\rm rem}$ is nonnegative, for either sign of
$\Delta_j$. Put $m_3=\operatorname{sech}^2(B_{\rm fit}+1)>0$.
On $S_k$, once $D_n\leq1$, the integrand has $q'''\geq m_3$ and
$a^{\rm rem}_j\geq m_3\Delta_j^2/2$. Thus
\[
\|\mathcal N_k(v)-z\|
\geq\frac{\|a^{\rm rem}\|}{H}
\geq\frac{m_3\|\Delta_{S_k}\|^2}{2H\sqrt p}
\geq\frac{c_*e^2}{\sqrt n},\qquad
c_*=\frac{m_3}{32H^3(1+M^2)^2}.
\]
This uses $p\leq n$ and the Hessian norm bound, without a
sign-preserving inverse assumption. Conversely, $q'''\leq1$ and
$\|\Delta^{\circ2}\|\leq\|\Delta\|_\infty\|\Delta\|$ give the
upper bound with $A_n=(M/2)S_nD_n=O(\log(en)^2/\sqrt n)$.

Let $d_v=H_k(v)^{-1}g_k(v)$. The identity
\[
H_k(v)^{-1}x_k=w+H_k(v)^{-1}[H_0-H_k(v)]w
\]
and the diagonal bound $|(H_0-H_k(v))_{jj}|\leq|\Delta_j|$ imply
\begin{equation}\label{eq:sharp-direction-infinity}
\|d_v\|_\infty\leq\rho_n:=S_nh_n(1+MS_n)
=O(\log(en)^{3/2}/\sqrt n).
\end{equation}
For $u=v+a$, direct subtraction of the Newton maps gives
\begin{equation}\label{eq:sharp-map-identity}
\begin{split}
\mathcal N_k(u)-\mathcal N_k(v)
={}&H_k(u)^{-1}\int_0^1[H_k(u)-H_k(v+ta)]a\,dt\\
&+H_k(u)^{-1}[H_k(u)-H_k(v)]d_v.
\end{split}
\end{equation}
The first term has norm at most $\|a^{\circ2}\|/2\leq\|a\|^2/2$;
the second at most $\|d_v\|_\infty\|a\|$ because each diagonal
entry changes by at most $|a_j|$.
This proves \eqref{eq:sharp-map-bound} globally in $a$.
\end{proof}

\begin{lemma}[The actual cap-one hidden recursion]
\label{lem:sharp-hidden-induction}
Use the hidden capped reference with $J=1$ and the unchanged $\tau_n^{(2)}$.
On $E_n$, put $E_H=4A_n+\tau_n^{(2)}$. For all sufficiently large $n$,
\begin{equation}\label{eq:sharp-hidden-centers}
\max_{k\leq T}\|C_k^H-b_k\|\leq E_H,\qquad
|e_k|\geq1\ \Longrightarrow\quad
\|C_k^H-b_k\|\geq c_*/(2\sqrt n).
\end{equation}
\end{lemma}
\begin{proof}
The prescribed $\tau_n^{(2)}=o(n^{-1/2})$, so eventually
$\rho_n+E_H/2\leq1/2$.
If the preceding actual center has error at most $E_H$, then
\eqref{eq:sharp-actual-cg}, \eqref{eq:sharp-newton-bounds} and
\eqref{eq:sharp-map-bound} give
\[
\|C_k^H-b_k\|\leq A_n+\rho_nE_H+E_H^2/2+\tau_n^{(2)}/8
\leq3A_n+5\tau_n^{(2)}/8\leq E_H.
\]
Exact initialization starts this induction at zero error.
It includes centers retained after a failed gate because the capped
algorithm retains the last center. If $|e_k|\geq1$, reverse triangle
inequalities give
\[
\|C_k^H-b_k\|\geq c_*/\sqrt n
             -\rho_nE_H-E_H^2/2-\tau_n^{(2)}/8
\geq c_*/(2\sqrt n),
\]
since the subtracted terms are $\widetilde O(n^{-1})$.
This is a bound for the actual inherited state and actual CG output.
\end{proof}

\subsection{Hidden path and first public marginal separation}
\label{app:sharp-first-two}
\begin{proposition}[Parts (i) and (ii)]
\label{prop:sharp-hidden-public-first}
For the model \eqref{eq:sharp-hidden-model}, there are
$E_n^H,E_n^P\subseteq\mathcal G_n$ with data probabilities tending
to one on which the actual comparisons in parts (i) and (ii),
respectively, fail for all sufficiently large $n$.
\end{proposition}
\begin{proof}
For the hidden path, let
$I_k=\mathbf1\{|y_k-x_k^\top b_k|\geq1\}$ and let
$\mathcal F_{k+1}$ contain the pairs with indices greater than $k$.
Given this field and $x_k$, at least one of the two fair response
signs makes the residual magnitude at least one. Therefore, for $t>0$,
\[
\mathbb E[e^{-tI_k}\mid\mathcal F_{k+1}]
\leq(1+e^{-t})/2.
\]
All $I_j$ with $j>k$ are $\mathcal F_{k+1}$-measurable.
Successively conditioning starting at $k=1$ gives
\[
\mathbb E\exp\!\left(-t\sum_{k\leq T}I_k\right)
\leq[(1+e^{-t})/2]^T.
\]
At $t=\log3$, Markov's inequality yields
\begin{equation}\label{eq:sharp-hidden-count}
\mathbb P_D\!\left(\sum_{k\leq T}I_k<T/4\right)
\leq e^{-c_{\rm count}T},\qquad
c_{\rm count}=\log(3/2)-(\log3)/4>0.
\end{equation}
This reverse conditioning does not require independent indicators.
Set $E_n^H=E_n\cap\{\sum_{k\leq T}I_k\geq T/4\}$.
By \eqref{eq:sharp-hidden-data-probability} and
\eqref{eq:sharp-hidden-count}, its data probability tends to one.
For fixed $D\in E_n^H$, the actual hidden path and independent
chronological retraining path are Gaussians with covariance
$\sigma_H^2I_{pT}$. Lemma~\ref{lem:sharp-hidden-induction} gives
\begin{equation}\label{eq:sharp-hidden-separation}
s_H^2:=\frac{\sum_{k\leq T}\|C_k^H-b_k\|^2}{\sigma_H^2}
\geq\frac{c_*^2\epsilon^2}{128n(\tau_n^{(2)})^2\log(1/\delta)}
\longrightarrow\infty.
\end{equation}
Indeed $n(\tau_n^{(2)})^2\leq C_\tau^2\log(en)^{2K_\tau}/n\to0$.
Lemma~\ref{lem:sharp-gaussian-shift} now makes both directed
divergences tend uniformly to one, proving part (i).
The $T/4$ center errors also exceed $\tau_n^{(2)}$ and force failed gates
by \eqref{eq:sharp-gate-lower}; the certification conclusion follows
from the law comparison, not from these gates.

For the first public release, a shrinking residual threshold is needed.
Conditional on the retained pairs, $x_1^\top b_1$ is $N(0,v)$ for
some $v\geq0$, independent of the response sign. If $v=0$, the
residual magnitude is one. For $v>0$ and $u\ne0$,
\[
\sup_{v>0}\frac{e^{-u^2/(2v)}}{\sqrt{2\pi v}}
=\frac1{\sqrt{2\pi e}|u|}.
\]
For $0<t\leq1/2$, each radius-$t$ interval about $+1$ or $-1$
has length $2t$ and lies in $|u|\geq1/2$. Its probability is at most
$4t/\sqrt{2\pi e}<2t$. Averaging the two signs shows
\begin{equation}\label{eq:sharp-first-smallball}
\mathbb P_D(|y_1-x_1^\top b_1|<t\mid\mathcal F_2)\leq2t.
\end{equation}
Put $t_n=n^{-1/8}$ and $E_n^P=E_n\cap\{|e_1|\geq t_n\}$.
Its failure probability is at most
$\mathbb P_D(E_n^c)+2n^{-1/8}\to0$.
The first actual public input is exactly $b_0$, and
\eqref{eq:sharp-newton-bounds} and \eqref{eq:sharp-actual-cg} give
\begin{equation}\label{eq:sharp-first-center}
\|C_1^P-b_1\|\geq c_*n^{-3/4}-\tau_n^{(2)}/8
\geq(c_*/2)n^{-3/4}
\end{equation}
eventually. Thus its true first marginal is
$N(C_1^P,\sigma_P^2I)$, with deterministic center given data, and
\begin{equation}\label{eq:sharp-first-separation}
s_P^2:=\frac{\|C_1^P-b_1\|^2}{\sigma_P^2}
\geq\frac{c_*^2\epsilon^2}{32\log(2/\delta)}
       \frac{n^{1/2}}{(n\tau_n^{(2)})^2}\longrightarrow\infty.
\end{equation}
Lemma~\ref{lem:sharp-gaussian-shift} proves part (ii).
There is no preceding publication-noise feedback at this first
release. The shrinking threshold changes only the proof's data
event, not the model or the deletion index.
\end{proof}

\subsection{A public-feedback model and its data geometry}
\label{app:sharp-public-model}
For part (iii), fix $\theta\in(0,1)$, for example $1/2$, and $a=1$.
For $p\geq2$, let
\begin{equation}\label{eq:sharp-public-model}
\begin{split}
v_p&=(0,1,\ldots,1)^\top/\sqrt{p-1},\qquad
\Sigma_p=I+\theta(e_1v_p^\top+v_pe_1^\top),\\
x_i&\stackrel{\rm iid}{\sim}N(0,\Sigma_p/p),\qquad y_i=0.
\end{split}
\end{equation}
Use squared loss, $\lambda=1$, and
\begin{equation}\label{eq:sharp-public-penalty}
r(\beta)=\sum_jq(\beta_j),\qquad
q''(t)=d+a\tanh(bt),\qquad
q'(t)=dt+(a/b)\log\cosh(bt),\qquad q(0)=0.
\end{equation}
We select $d>a$ after a fixed spectral envelope and then select a
finite $b>0$, before observing any data or mechanism noise.
The covariance eigenvalues are $1-\theta,1+\theta,1,\ldots,1$.
The penalty satisfies $q'(0)=0$, $d-a\leq q''\leq d+a$,
$q(t)\geq(d-a)t^2/2$, and
\[
\|\nabla^2r(u)-\nabla^2r(v)\|_{\mathrm F}^2
\leq a^2b^2\|u-v\|^2.
\]
Thus $\mu=d-a>0$, the global Hessian bound is $d+a$, and its
Frobenius Lipschitz constant is the fixed number $ab$.
Squared loss has the required smoothness and quadratic growth;
zero responses satisfy all tail and moment requirements. With an
independent test pair from this same population, all of
Assumptions~\ref{ass:design}--\ref{ass:Penalty} hold. Every exact
fit, including $b_0$, is zero, by normalization and strong convexity.

Write $A_k=X_k^\top X_k$ and $H_{0k}=A_k+dI$.
Cancellation of the empirical quadratic part in $H_k(u)u-g_k(u)$ gives
\begin{equation}\label{eq:sharp-public-newton}
H_k(u)=A_k+\operatorname{diag}(d+a\tanh(bu_j)),
\qquad \mathcal N_k(u)=H_k(u)^{-1}w(u),
\end{equation}
where $w$ acts coordinatewise and
\[
w(t)=a\{t\tanh(bt)-b^{-1}\log\cosh(bt)\}
     =\int_0^t ab\,s\operatorname{sech}^2(bs)\,ds.
\]
The function is even and nonnegative. Direct integration gives
\begin{equation}\label{eq:sharp-w-bounds}
\begin{aligned}
0\leq w(t)&\leq abt^2/2 &&(t\in\mathbb R),\\
w(t)&\geq\tfrac12ab\operatorname{sech}^2(1)t^2 &&(|bt|\leq1),\\
|w(t)-(ab/2)t^2|&\leq ab^3|t|^4/4 &&(t\in\mathbb R).
\end{aligned}
\end{equation}
The last inequality follows by integrating
$\tanh^2(bs)\leq b^2s^2$, including negative $t$ by evenness.

\begin{lemma}[Data geometry for the feedback model]
\label{lem:sharp-public-data}
There is a fixed $M_A<\infty$ such that, with
$r_0=\theta(1-c')/(4\gamma_+)>0$, the data event
\begin{equation}\label{eq:sharp-public-data-event}
\mathcal H_n=
\{\|A_k\|\leq M_A,\quad
e_1^\top A_k\mathbf1\geq r_0\sqrt p\quad(1\leq k\leq T)\}
\end{equation}
has probability $1-O(n^{-m})$ for each fixed $m>0$.
The event and model constants can be chosen independently of $m$.
After choosing
\begin{equation}\label{eq:sharp-public-d-choice}
d>a,\qquad d\geq4M_A^2/r_0,\qquad
c_H=r_0/(2d^2),\qquad H_{\max}=M_A+d+a,
\end{equation}
one has, on $\mathcal H_n$ and eventually for all $k\leq T$,
\begin{equation}\label{eq:sharp-inverse-row-sum}
e_1^\top H_{0k}^{-1}\mathbf1\leq-c_H\sqrt p.
\end{equation}
All actual Hessians have spectrum in $[\mu,H_{\max}]$, globally in $u$.
\end{lemma}
\begin{proof}
Use the two sphere nets constructed in the proof of
Lemma~\ref{lem:sharp-hidden-geometry}.
Here a fixed design bilinear form has Gaussian variance at most
$(1+\theta)/p$, so
\[
\mathbb P_D(\|X\|>B)
\leq2\exp\{(n+p)\log9-pB^2/[8(1+\theta)]\}.
\]
A sufficiently large fixed $B$, using $p/n\geq\gamma_-$, makes
this at most $2e^{-cn}$ for a fixed $c>0$.
Set $M_A=B^2$; every retained covariance obeys
$\|A_k\|\leq\|X\|^2$.
For the row sum, direct covariance calculation gives, uniformly in $k$,
\[
\mathbb E_D[e_1^\top A_k\mathbf1]
=\frac{(n-k)(1+\theta\sqrt{p-1})}{p}
\geq2r_0\sqrt p
\]
eventually. Let $W_i=(\sqrt p\,x_{i1})(x_i^\top\mathbf1)$.
The two Gaussian factors have variances $1$ and
$1+2\theta\sqrt{p-1}/p$, respectively. Their variances and all fixed
absolute moments of $W_i$ are uniformly bounded, by Gaussian moments
and Cauchy--Schwarz. In the $2q$-th centered moment of a sum of the
independent $W_i$, every term with a singleton index vanishes.
All remaining terms have at most $q$ distinct indices, so expansion
and the moment bounds give
\[
\mathbb E_D\left|\sum_{i>k}(W_i-\mathbb E_DW_i)\right|^{2q}
\leq C_q n^q.
\]
A row-sum deviation of $r_0\sqrt p$ corresponds to a $W$-sum
deviation of $r_0p$. Markov's inequality therefore gives a failure
of $O(n^{-q})$ for one prefix and $O(n^{1-q})$ after the union over
$T$ prefixes. For a prescribed $m$, take $q>m+1$.
This changes proof constants, not the data event or the model.

On $\mathcal H_n$, the exact identity
\[
(dI+A_k)^{-1}=d^{-1}I-d^{-2}A_k
                  +d^{-2}A_k^2(dI+A_k)^{-1}
\]
implies
\[
e_1^\top H_{0k}^{-1}\mathbf1
\leq d^{-1}-r_0d^{-2}\sqrt p+M_A^2d^{-3}\sqrt p
\leq-c_H\sqrt p.
\]
The remainder costs at most one quarter of the negative term by
\eqref{eq:sharp-public-d-choice}; eventually the fixed $d^{-1}$
costs another quarter. The global spectrum bound follows from
$A_k\succeq0$, $\|A_k\|\leq M_A$ and $d-a\leq q''\leq d+a$.
This argument uses a scalar row-sum estimate and a spectral upper
bound; it does not approximate the sample covariance by its mean in
operator norm.
\end{proof}

\subsection{Original innovations and the first actual correction}
\label{app:sharp-public-innovations}
Fix $D\in\mathcal H_n$ and retain
\eqref{eq:sharp-public-calibration}. All probabilities in this
subsection are conditional mechanism probabilities for this fixed
dataset. Define
\begin{equation}\label{eq:sharp-noise-constants}
\begin{split}
m_Z&=\sqrt{\kappa_\sigma t_-\gamma_-/2},\qquad
M_Z=\sqrt{2\kappa_\sigma c'\gamma_+},\\
D_Z&=2\kappa_\sigma c'\sqrt{\gamma_+},\qquad
r_n=C_m\sqrt{\log(en)/n}.
\end{split}
\end{equation}

\begin{lemma}[A common event of the original Gaussian innovations]
\label{lem:sharp-public-noise}
For each fixed $m>0$, $C_m$ can be chosen so that an event
$\mathcal E_n$ of the original innovations has probability
$1-O(n^{-m})$, uniformly over $D\in\mathcal H_n$, and satisfies
\begin{equation}\label{eq:sharp-noise-event}
m_Z\leq\|Z_j\|\leq M_Z,\qquad
\|Z_j\|_\infty\leq r_n,\qquad
\|Z_j^{\circ2}\|\leq D_Z/\sqrt n
\quad(1\leq j\leq T).
\end{equation}
For $h_k^\top=e_1^\top H_{0k}^{-1}$, the same event also satisfies
\begin{equation}\label{eq:sharp-signed-square-event}
|h_k^\top(Z_{k-1}^{\circ2}-\sigma^2\mathbf1)|
\leq(c_H/2)\sigma^2\sqrt p\qquad(2\leq k\leq T).
\end{equation}
\end{lemma}
\begin{proof}
The mean $p\sigma^2$ lies between
$\kappa_\sigma t_-\gamma_-$ and $\kappa_\sigma c'\gamma_+$.
The identity $\mathbb E e^{t\chi_p^2}=(1-2t)^{-p/2}$ for $t<1/2$
and Chernoff's inequality give
$\chi_p^2\in[p/2,2p]$ with failure at most $2e^{-cp}$ for an
absolute $c>0$. Applying this to $\|Z_j\|^2/\sigma^2$ and taking
a union over $j$ gives the norm bounds. Scalar Gaussian tails and
a union over at most $Tp$ coordinates give the maximum-coordinate
bound for sufficiently large $C_m$.

For a standard Gaussian vector $G$, apply the centered-moment
expansion from Lemma~\ref{lem:sharp-public-data} to the independent
variables $G_i^4-3$. Their fixed moments are finite, hence
\[
\mathbb E\left|\sum_{i=1}^p(G_i^4-3)\right|^{2q}\leq C_qp^q.
\]
Markov bounds $\mathbb P(\sum_iG_i^4>4p)$ by $C_qp^{-q}$.
Choosing $q>m+1$ and taking a union over $T$ yields the required
polynomial confidence. Since $Z=\sigma G$, this gives
\[
\|Z^{\circ2}\|\leq2\sigma^2\sqrt p\leq D_Z/\sqrt n.
\]

The rows $h_k$ are deterministic given data and satisfy
$\|h_k\|_2,\|h_k\|_\infty\leq1/d$.
For any signed weights $h$ and independent standard Gaussian
coordinates, direct one-dimensional integration gives
\begin{equation}\label{eq:sharp-signed-square-mgf}
\begin{split}
\mathbb E\exp\!\left(t\sum_jh_j(G_j^2-1)\right)
&=\prod_j e^{-th_j}(1-2th_j)^{-1/2}\\
&\leq\exp(2t^2\|h\|_2^2),
\qquad |t|\leq(4\|h\|_\infty)^{-1}.
\end{split}
\end{equation}
Indeed, for $|x|\leq1/4$, the logarithmic series gives
\[
-x-\tfrac12\log(1-2x)
\leq\sum_{r\geq2}\frac{2^{r-1}|x|^r}{r}
\leq\frac{x^2}{1-2|x|}\leq2x^2.
\]
This covers both signs of $t$ and every weight.
For a deviation $u>0$, take the smaller of
$u/(4\|h\|_2^2)$ and $1/(4\|h\|_\infty)$ as the positive
Chernoff parameter and use its negative for the other tail. This gives
\[
\mathbb P\!\left(\left|\sum_jh_j(G_j^2-1)\right|>u\right)
\leq2\exp\!\left[-c\min\left\{
\frac{u^2}{\|h\|_2^2},\frac{u}{\|h\|_\infty}\right\}\right]
\]
for an absolute $c>0$.
Take $u=c_H\sqrt p/2$ and union over $k$ to obtain a total failure
$O(ne^{-c_0\sqrt p})$, for a fixed positive constant $c_0$ in this
tail bound. Combining the events proves the lemma. Only events of
the original innovations have been used; the actual output law is
not replaced by a law conditioned on their success.
\end{proof}

\begin{lemma}[Uniform first correction]
\label{lem:sharp-public-first-correction}
Fix a finite $K$. On $\mathcal H_n$ and $\mathcal E_n$, suppose the
preceding actual center $c$ satisfies $\|c\|\leq K/n$ and the input
at deletion $k\geq2$ is $u=c+Z_{k-1}$. Let $u_1$ be its first
actual Newton--CG correction. With
\begin{equation}\label{eq:sharp-first-correction-constants}
s_*=\kappa_\sigma t_-\sqrt{\gamma_-},\qquad
\ell_b=ab c_Hs_*/8,
\end{equation}
one has eventually $(u_1)_1\leq-\ell_b/\sqrt n$, uniformly over
the entire preceding-center ball and all $k\geq2$.
\end{lemma}
\begin{proof}
Abbreviate $Z=Z_{k-1}$ and set
\[
s=r_n+K/n,\qquad F=2r_nK/n+K^2/n^2,\qquad
Q=D_Z/\sqrt n+F.
\]
Coordinatewise algebra gives
$\|u^{\circ2}-Z^{\circ2}\|\leq F$,
$\|u\|_\infty\leq s$, and $\|u^{\circ2}\|\leq Q$.
Add and subtract $H_{0k}^{-1}w(u)$ and
$(ab/2)H_{0k}^{-1}u^{\circ2}$ in the exact Newton map.
The inverse perturbation has norm at most $ab s/(d\mu)$, since
$\|H_k(u)-H_{0k}\|\leq ab s$. Combining this with
\eqref{eq:sharp-w-bounds} and the actual CG bound yields
\begin{equation}\label{eq:sharp-first-correction-expansion}
\begin{split}
\left\|u_1-\frac{ab}{2}H_{0k}^{-1}Z^{\circ2}\right\|
&\leq\frac{ab}{2d}F+\frac{ab^3}{4d}s^2Q
       +\frac{a^2b^2}{2d\mu}sQ+\frac1{8n}\\
&=o(n^{-1/2}).
\end{split}
\end{equation}
The four terms respectively control the coordinate-square difference,
the fourth-order remainder, the inverse perturbation times
$\|w(u)\|\leq abQ/2$, and the actual CG displacement.
All bounds hold uniformly on the ball, so they do not require $c$
to be independent of $Z$. The limit is taken after fixing $b$ and $K$.

By \eqref{eq:sharp-inverse-row-sum} and
\eqref{eq:sharp-signed-square-event},
\[
h_k^\top Z^{\circ2}\leq-(c_H/2)\sigma^2\sqrt p.
\]
As $\sigma^2\sqrt p\geq s_*/\sqrt n$, the first coordinate of
$(ab/2)H_{0k}^{-1}Z^{\circ2}$ is at most
$-ab c_Hs_*/(4\sqrt n)=-2\ell_b/\sqrt n$.
Absorbing \eqref{eq:sharp-first-correction-expansion} proves the claim.
\end{proof}

\subsection{Two actual public updates}
\label{app:sharp-public-two}
Define
\begin{equation}\label{eq:sharp-public-recursion-constants}
\begin{split}
A_b&=abD_Z/\mu+1,\qquad K_b=1+\frac{ab}{2\mu}A_b^2,\\
L_b&=\frac{ab\operatorname{sech}^2(1)}{2H_{\max}}\ell_b^2-\frac18
     =C_*b^3-\frac18,\\
C_*&=\frac{a^3\operatorname{sech}^2(1)c_H^2s_*^2}{128H_{\max}}>0.
\end{split}
\end{equation}
The constant $C_*$ is independent of $b$. Choose once and for all
\begin{equation}\label{eq:sharp-public-b-choice}
b>\left[\frac{1/8+\max\{2,\sqrt{128\kappa_\sigma}\}}{C_*}
  \right]^{1/3}.
\end{equation}
Then $L_b>1$ and $L_b^2/(2\kappa_\sigma)\geq64$.
The choice order is $\theta,a,M_A,r_0,d$, then $b$, then $A_b,K_b$,
and finally sufficiently large $n$. In particular, large but fixed
$b$ is allowed by the original regularity assumptions, and
$H_{\max}$ is independent of $b$.

\begin{proposition}[Public cap two under the prescribed calibration]
\label{prop:sharp-public-two}
For the fixed model \eqref{eq:sharp-public-model}--\eqref{eq:sharp-public-penalty}
with the preceding choices, let $P_D^{(2)}$ denote the actual public
transcript law with $J=2$ and let
$Q_D=\bigotimes_{k=1}^T N(0,\sigma^2I_p)$ be its independent
retraining reference. Then
\begin{equation}\label{eq:sharp-public-two-margin}
\liminf_{n\to\infty}\inf_{D\in\mathcal H_n}
\mathrm D_\epsilon(P_D^{(2)},Q_D)
\geq(1-e^{-15})(1-e^{-2})>0.86>\delta.
\end{equation}
\end{proposition}
\begin{proof}
Work on the original innovation event $\mathcal E_n$.
We induct on the actual final centers with invariant
$\|C_k\|\leq K_b/n$. The first input is the exact fit zero, so
the zero-gradient convention leaves it zero and the first gate
passes: $C_1=0$.
For $k\geq2$, assume the invariant for $C_{k-1}$. Then
\[
\|(C_{k-1}+Z_{k-1})^{\circ2}\|
\leq D_Z/\sqrt n+2r_nK_b/n+K_b^2/n^2
\leq2D_Z/\sqrt n
\]
eventually. Equations~\eqref{eq:sharp-w-bounds} and
\eqref{eq:sharp-actual-cg} give
$\|u_1\|\leq A_b/\sqrt n$ for the first actual correction.
Lemma~\ref{lem:sharp-public-first-correction}, with $K=K_b$, gives
$\|u_1\|\geq\ell_b/\sqrt n$.
Therefore the first actual gate fails:
\[
\|g_k(u_1)\|\geq\mu\ell_b/\sqrt n>\mu/n.
\]
The algorithm consequently does execute its second correction.
For its actual final center,
\[
\|C_k\|\leq\frac{ab}{2\mu}\|u_1^{\circ2}\|+\frac1{8n}
\leq\frac{abA_b^2}{2\mu n}+\frac1{8n}\leq K_b/n.
\]
This closes the invariant without assuming that the final gate passes.

Once $bA_b/\sqrt n\leq1$, the lower bound in
\eqref{eq:sharp-w-bounds} applies to every coordinate of $u_1$.
Its first coordinate and the global Hessian bound imply
\[
\|\mathcal N_k(u_1)\|\geq\frac{\|w(u_1)\|}{H_{\max}}
\geq\frac{ab\operatorname{sech}^2(1)\ell_b^2}{2H_{\max}n}.
\]
Subtracting the actual, same-order CG error $1/(8n)$ gives
$\|C_k\|\geq L_b/n$. Since $L_b>1$, the second gate also fails
by \eqref{eq:sharp-gate-lower}. The cap is exhausted and the actual
algorithm still publishes $C_k+Z_k$.
Thus both lower bounds concern the actual capped recursion.

All exact fits are zero. On $\mathcal E_n$, the actual predictable
energy from Lemma~\ref{lem:sharp-adapted-test} satisfies
\begin{equation}\label{eq:sharp-public-two-energy}
V_T\geq\frac{(T-1)L_b^2}{n^2\sigma^2}
    =\frac{T-1}{T}\frac{L_b^2}{\kappa_\sigma}\geq64
\end{equation}
eventually. This event has probability tending uniformly to one
over fixed $D\in\mathcal H_n$ by Lemma~\ref{lem:sharp-public-noise}.
Taking $v=64$ in \eqref{eq:sharp-adapted-test} yields
\[
\liminf_{n\to\infty}\inf_{D\in\mathcal H_n}
\mathrm D_\epsilon(P_D^{(2)},Q_D)
\geq(1-e^{\epsilon-16})(1-e^{-2})
\geq(1-e^{-15})(1-e^{-2}),
\]
which proves \eqref{eq:sharp-public-two-margin}.
This is a fixed strict separation margin. A forward divergence
exceeding $\delta$ suffices to rule out bilateral certification.
\end{proof}

\subsection{One actual public update and completion}
\label{app:sharp-public-one}
\begin{proposition}[Public cap one under the same calibration]
\label{prop:sharp-public-one}
Keep the same fixed model, data class, and calibration, but run the
public capped reference with $J=1$. Its transcript law $P_D^{(1)}$
satisfies
\[
\inf_{D\in\mathcal H_n}\mathrm D_\epsilon(P_D^{(1)},Q_D)
\longrightarrow1.
\]
\end{proposition}
\begin{proof}
Choose a fixed $A_b'>abD_Z/(2\mu)+1$.
On $\mathcal E_n$, the invariant is now
$\|C_k\|\leq A_b'/\sqrt n$. It starts at $C_1=0$.
If the preceding center $c$ satisfies the invariant, then
\[
\|(Z+c)^{\circ2}\|
\leq D_Z/\sqrt n+2r_nA_b'/\sqrt n+(A_b')^2/n
=D_Z/\sqrt n+o(n^{-1/2}).
\]
Using \eqref{eq:sharp-public-newton}, \eqref{eq:sharp-w-bounds}
and \eqref{eq:sharp-actual-cg}, the next actual center therefore
has norm at most $A_b'/\sqrt n$ eventually, closing this separate
cap-one induction.

For every $k\geq2$, the input $u=C_{k-1}+Z_{k-1}$ also satisfies
$\|u\|\geq m_Z/2$ and
$\|u\|_\infty\leq r_n+A_b'/\sqrt n\to0$.
Since $b$ is fixed, the coordinatewise lower bound in
\eqref{eq:sharp-w-bounds} applies eventually.
Cauchy--Schwarz gives $\|u^{\circ2}\|\geq\|u\|^2/\sqrt p$,
and hence
\begin{equation}\label{eq:sharp-public-one-center}
\begin{split}
\|\mathcal N_k(u)\|
&\geq\frac{ab\operatorname{sech}^2(1)}{2H_{\max}\sqrt p}\|u\|^2
\geq c_b/\sqrt n,\\
c_b&=\frac{ab\operatorname{sech}^2(1)m_Z^2}
             {8H_{\max}\sqrt{\gamma_+}}>0.
\end{split}
\end{equation}
Subtracting $1/(8n)$ for actual CG gives
$\|C_k\|\geq c_b'/\sqrt n$ eventually, with $c_b'=c_b/2$.
The gate fails but the last center is still published.
The energy of this actual cap-one process thus obeys
\begin{equation}\label{eq:sharp-public-one-energy}
V_T\geq\frac{(T-1)(c_b')^2}{n\sigma^2}
    =\frac{T-1}{T}\frac{(c_b')^2n}{\kappa_\sigma}
\longrightarrow\infty
\end{equation}
on the original innovation event, whose conditional probability
tends uniformly to one. Apply \eqref{eq:sharp-adapted-test} first
at each fixed $v$, take a limit inferior uniformly in $D$, and then
let $v\to\infty$. This proves the stated forward divergence limit
for the separate cap-one recursion.
\end{proof}

\begin{proof}[Proof of Theorem~\ref{thm:prescribed-update-sharpness}]
Proposition~\ref{prop:sharp-hidden-public-first} proves (i) and (ii)
on $E_n^H$ and $E_n^P$, respectively. For (iii), take
$E_n^{PP}=\mathcal H_n\cap\mathcal G_n$.
Lemma~\ref{lem:sharp-public-data} and the existing data-class bound
give $\mathbb P_D(E_n^{PP})\to1$.
Propositions~\ref{prop:sharp-public-two} and
\ref{prop:sharp-public-one} prove eventual fixed-data failure on
this same event for each cap. The former gives a forward divergence
with a limit inferior strictly above $\delta$; the latter gives a
forward divergence tending to one. Each suffices to violate a
bilateral comparison. Both comparisons use the complete actual
output law, including histories outside $\mathcal E_n$.

All bounds are uniform over their respective data events.
Taking the maximum of the finitely many sufficiently-large-$n$
thresholds gives the claimed quantifiers. The two model families
and their constants are fixed before data and mechanism noise;
the second family's $b$ may depend on the fixed privacy and
dimension/horizon bounds. The full deterministic envelope in
\eqref{eq:sharp-two-calibrations} is retained, and the threshold may
depend on its fixed constants and logarithmic power.
This proves the stated capped-reference obstructions. The next
subsection establishes stopping-count lower bounds for the guarded
algorithms.
\end{proof}
\subsection{Stopping-count lower bounds for the guarded algorithms}
\label{app:stopping-sharpness}
\begin{proof}[Proof of Theorem~\ref{thm:stopping-sharpness}]
Use the corresponding model families, tolerances, noise scales and
high-probability data events constructed above. Couple capped and
guarded processes using the same original Gaussian innovations.

For pathwise partial unlearning, Lemma~\ref{lem:sharp-hidden-induction}
and \eqref{eq:sharp-hidden-count} imply that the cap-one hidden path
has a failed gate on $E_n^H$: at a residual-index round its center
error is at least $c_*/(2\sqrt n)>\tau_n^{(2)}$. Before the first failed
one-update gate, both processes retained the same centers. Thus at
that first round the guarded process also fails its first gate and
requires $N_k>1$. Since $B_{\mathrm N}=J_\star=2$, this failed first gate
is followed by a second Newton correction before recovery is possible.
This transfers existence of a failed round, not the number of later
failures along the now-different paths.

For pointwise perfect unlearning, both first-round inputs are exactly
$b_0$. On $E_n^P$, \eqref{eq:sharp-first-center} gives error at least
$(c_*/2)n^{-3/4}>\tau_n^{(2)}$ after the first actual CG correction. Hence
$N_1>1$, independently of subsequent publication noise. The budget
$B_{\mathrm N}=J_\star=2$ permits the second correction before any
recovery.

For pathwise perfect unlearning, all exact fits in the feedback model
are zero. The first correction leaves zero and passes its gate in
both processes, so the second deletion starts at the same $Z_1$.
On $\mathcal H_n\cap\mathcal G_n$ and the original innovation event
$\mathcal E_n$, Lemma~\ref{lem:sharp-public-first-correction} with
preceding center zero gives first-correction norm at least
$\ell_b/\sqrt n>1/n$. The second-correction lower bound in the proof
of Proposition~\ref{prop:sharp-public-two} gives norm at least
$L_b/n>1/n$. Strong convexity forces both gates to fail. With
$B_{\mathrm N}=J_\star=3$, these failures force the third Newton
correction before recovery can be invoked, hence $N_2>2$.
Lemma~\ref{lem:sharp-public-noise} gives conditional probability
$1-O(n^{-d})$ for every fixed $d>0$, uniformly on the stated data
class. Its data probability tends to one separately. These are
existential-model lower bounds with fixed calibrations, not claims
that all models or all rounds require the displayed counts.
\end{proof}

\section{Finite recovery and work of the guarded algorithms}
\label{app:recovery-proof}
This appendix specifies the recovery procedure in
Algorithm~\ref{alg:sequential-newton-cg}, proves its finite termination
at the prescribed center tolerance, and establishes the resulting
certification and work guarantees. All optimization claims use exact
arithmetic. Work is measured in the scalar-oracle cost model of
Appendix~\ref{app:cg-proof}, excluding initial training. Objective and
derivative evaluations are assumed jointly measurable in the data and
iterate, as required for the release kernels.

\subsection{Recovery procedure and finite termination}
\label{app:recovery-procedure}
Recovery is invoked only after $B_{\mathrm N}=J_\star$ Newton--CG
updates have been attempted at the current deletion and the gradient
gate checked after the last update remains unsatisfied. If an earlier
gate passes, the Newton prefix ends immediately. Recovery continues
optimization on the retained data until the same gradient gate passes.
This fallback covers datasets outside $\mathcal G_n$ and, in the
perfect setting, noise realizations for which the prefix tracking
guarantee does not apply even when $D\in\mathcal G_n$.
The algorithm checks the gradient gate rather than membership in
these probabilistic events.

We use gradient descent with Armijo backtracking. The line search
selects a descent step using objective evaluations, without requiring
a known global smoothness constant. The initialization below places
recovery in a sublevel set controlled by the retained data, independently
of the preceding release history. Together with strong convexity and
local smoothness, these properties yield finite termination at every
positive tolerance.

Fix a retained objective $F=F_{\backslash k}$, its minimizer $b$, and
$\tau=\tau_n>0$. Write $g=\nabla F$ and $\mu=\lambda\nu>0$.
By Lemma~\ref{lem:uniform_strong_convexity}, for every finite dataset,
$F(u)\geq\mu\|u\|^2/2$ and
$\|u-b\|\leq\|g(u)\|/\mu$.
Given a finite iterate $w$ returned by the Newton prefix, recovery starts at
\[
 s=\begin{cases}w,&F(w)\leq F(0),\\0,&F(w)>F(0).\end{cases}
\]
Thus $F(s)\leq F(0)$, while retaining the current iterate whenever
its objective value is no larger than the value at zero.
It returns immediately if $\|g(s)\|\leq\mu\tau$.
Otherwise, at each current iterate $u$, it first tries $a=a_0>0$.
Whenever the following condition fails, it replaces $a$ by $\beta_A a$
and tests again until the condition holds:
\begin{equation}\label{eq:recovery-armijo}
 F(u-a g(u))\leq F(u)-c_A a\|g(u)\|^2.
\end{equation}
It then sets $u\leftarrow u-a g(u)$ and retests the gradient gate.
For Algorithm~\ref{alg:sequential-newton-cg}, we take
$a_0=1$ and $c_A=\beta_A=1/2$. The analysis also permits any fixed
$a_0>0$ and $c_A,\beta_A\in(0,1)$.
The procedure continues until the gradient gate passes, without an
additional iteration or backtracking budget.

\begin{lemma}[Finite measurable recovery]
\label{lem:recovery-termination}
Under Assumptions~\ref{ass:loss} and~\ref{ass:Penalty}, the preceding recovery terminates
at a finite measurable center $c$ for every finite retained dataset,
every finite $w$, and every $\tau>0$, and
\[
 \|g(c)\|\leq\mu\tau,\qquad \|c-b\|\leq\tau.
\]
A fixed finite number of the prescribed Newton--CG updates,
followed when needed by this recovery, therefore defines a finite
measurable center at every finite release history.
\end{lemma}
\begin{proof}
Let $F_0=F(0)$. If $F_0=0$, nonnegativity and strong convexity force
$s=b=0$, so the initial gate passes. Otherwise put
$B_0=(2F_0/\mu)^{1/2}$.
The accepted recovery iterates remain in $\{F\leq F_0\}$ and hence
in the ball of radius $B_0$. This follows inductively from
\eqref{eq:recovery-armijo}, once a trial is accepted.

For this fixed finite dataset choose any finite nondecreasing Hessian
envelope $\mathcal L(R)\geq\mu$ on the ball of radius $R$; one exists
because $F$ is $C^2$. Define
\[
 G_0=\|g(0)\|+\mathcal L(B_0)B_0,\qquad
 R_0=B_0+a_0G_0,\qquad L_+=\mathcal L(R_0).
\]
For an accepted iterate $u$, integration from zero gives
$\|g(u)\|\leq G_0$. Thus every trial segment
$\{u-tg(u):0\leq t\leq a_0\}$ lies in the radius-$R_0$ ball,
including trials subsequently rejected. Taylor integration yields
\[
 F(u-a g(u))\leq F(u)-a\|g(u)\|^2
                         +\tfrac12 L_+a^2\|g(u)\|^2.
\]
Consequently every $a\leq a_*:=2(1-c_A)/L_+$ is accepted.
Each line search has at most $j_*+1$ trial evaluations, where
\[
 j_*=\max\left\{0,
 \left\lceil\frac{\log(a_0/a_*)}{\log(1/\beta_A)}\right\rceil\right\},
\]
and its accepted step is at least
$\underline a=\min\{a_0,\beta_A a_*\}>0$.
This proves the inductive sublevel assertion and makes the same
constants valid throughout recovery.

Each update taken while the gate fails decreases $F$ by more than
$c_A\underline a\mu^2\tau^2$. Since $F\geq0$, the number of such
updates is at most the conservative bound
\begin{equation}\label{eq:recovery-update-bound}
 N_* = 1+\left\lfloor
       \frac{F_0}{c_A\underline a\mu^2\tau^2}\right\rfloor.
\end{equation}
In particular the gate is reached after finitely many evaluations.
Strong convexity then gives the asserted center-distance bound.
Neither these envelope constants nor the unknown minimizer is needed
by the backtracking implementation.

For the Newton prefix, every Hessian is positive definite and exact
CG terminates within $p$ iterations, allowing a zero-iteration return
of the zero direction when the initial residual meets the absolute
tolerance. Thus a fixed finite prefix returns a finite point.
All finite arithmetic, oracle evaluations, and comparison branches
are measurable. For recovery, the sets specifying any finite list of
backtracking counts and a terminal iteration are measurable. The
returned point is the corresponding finite composition on each such
set. Their countable union covers the domain by finite termination,
so the returned map is measurable. This argument requires no good-data
or good-noise event.
\end{proof}

\subsection{Certification after recovery}
\label{app:recovery-certification}
The termination result holds for every finite dataset and finite
release history. It therefore supplies the center accuracy needed
for certification even when the fast prefix guarantee is unavailable.

\begin{corollary}[Certification after guarded recovery]
\label{cor:guarded-certification}
Suppose Algorithm~\ref{alg:sequential-newton-cg} preserves its respective state rules,
uses a finite Newton budget (in particular $B_{\mathrm N}=J_\star$),
and invokes the preceding recovery before publishing whenever the
budget is exhausted and the gradient gate has not passed.
Its actual centers satisfy $\|c_k(h)-b_k\|\leq\tau_n$ at every
finite history $h$. With fresh independent Gaussian publication noise,
Theorem~\ref{thm:gaussian-certification} applies directly to the actual
mechanism. In particular its Gaussian inequality gives pathwise
certification with $v=T\tau_n^2/\sigma_n^2$, and pointwise
certification with $v=\tau_n^2/\sigma_n^2$.
\end{corollary}
\begin{proof}
A center returned by either the Newton gate or recovery obeys the
same gradient certificate. Lemma~\ref{lem:recovery-termination}
supplies a finite measurable center also on histories where the
Newton prefix fails. The Gaussian conditional kernels therefore
satisfy every hypothesis of
Theorem~\ref{thm:gaussian-certification}; its bounded-center mixture
statement applies to the public-state pointwise marginals.
Only model releases are included in this comparison: iteration counts,
recovery indicators, and runtimes are not additional certified outputs.
\end{proof}
For $0<\epsilon\leq1$ and $0<\delta<1/2$, the pathwise calibration
$\sigma_n=2\tau_n\sqrt{2T\log(1/\delta)}/\epsilon$ gives
\[
 v/2+\sqrt{2v\log(1/\delta)}
 =\frac{\epsilon^2}{16\log(1/\delta)}+\frac\epsilon2<\epsilon.
\]
The same calculation with $T=1$ covers the pointwise calibration;
replacing $\log(1/\delta)$ by $\log(2/\delta)$ is conservative.
This argument uses the prescribed center tolerance, including $\tau_n^{(1)}$
in the one-update hidden-state configuration. Once Newton iteration or
recovery has returned a center that passes the gradient gate, the Gaussian
comparison applies with that same tolerance.

\subsection{Arithmetic work on good datasets}
\label{app:recovery-work}
We now fix $D\in\mathcal G_n$ and take expectations over the
publication noise. The comparison with $F(0)$ gives a polynomial
bound on each recovery, uniformly over preceding release histories.
Combining this bound with the prefix tracking event controls the
expected work of the guarded algorithm.

\begin{proposition}[Guarded arithmetic work]
\label{prop:guarded-work}
Fix the model and data-confidence constants, $a_0,c_A,\beta_A$, and
$\mu>0$. Let $p\asymp n$, $T\leq\lfloor c'n\rfloor$,
$\tau_n\geq n^{-1}$, and let the maximum Newton budget be
$B_{\mathrm N}=J_\star$, where $J_\star\in\{1,2,3\}$ is fixed.
The gradient gate is checked after every update, including update
$B_{\mathrm N}$, before recovery is considered.
On $\mathcal G_n$, there is a deterministic per-recovery work bound
\[
 R_{\mathrm{work},n}\leq C_R n^{A_R}\log(en)^{b_R}
\]
for finite constants independent of the release history and of the
chosen mechanism-noise confidence exponent.
For a fixed $D\in\mathcal G_n$, suppose the corresponding prefix
tracking theorem supplies an event $E$ such that
$\mathbb P_D(E^c)\leq q_n$ and, on $E$, every deletion passes its gate
within $J_\star$ corrections and all called Hessians satisfy the
stated fast CG work bounds. Then the guarded algorithm has no recovery
on $E$, the probability of any recovery is at most $q_n$, and
\begin{equation}\label{eq:guarded-expected-work}
 \mathbb E_D W
 \leq W_{\mathrm{fast},n}
       +q_n\{C_B B_{\mathrm N}Tnp^2+T R_{\mathrm{work},n}\},
\end{equation}
where $W_{\mathrm{fast},n}=\widetilde O(Tnp)$ bounds its work on $E$.
In particular, if for every fixed $d>0$ the tracking theorem supplies
such an event $E=E_d$ with $q_n\leq n^{-d}$ for all sufficiently
large $n$ (allowing the fast-work constant to depend on $d$), then
$\mathbb E_D W=\widetilde O(Tnp)$ uniformly over fixed
$D\in\mathcal G_n$. If $E$ is deterministic on $\mathcal G_n$,
no recovery cost is incurred there.
\end{proposition}
\begin{proof}
We first bound recovery without restricting the preceding history.
On $\mathcal G_n$, all prefixes satisfy
$\|X_k\|_{\rm op}\leq M=O(1)$ and
$\max_i|y_i|\leq C_y(n)=\operatorname{polylog}(n)$.
The warm/zero comparison gives
\[
 F_0\leq Cn(1+C_y(n)^s),\qquad
 B_0=O(\sqrt n\operatorname{polylog}(n)),\qquad
 \|g(0)\|\leq CM\sqrt n(1+C_y(n)^s).
\]
Here $r(0)=0$ and $\nabla r(0)=0$ are used.
To control the Hessian on the enlarged trial ball, the primitive
loss bounds imply, for every $r\geq0$,
\begin{equation}\label{eq:recovery-loss-hessian}
 \sup_{|t|\leq r}\ddot\ell(y,t)
 \leq C_s(1+|y|^s)(1+r^{s+1}).
\end{equation}
Indeed Taylor integration between zero and one bounds
$\ddot\ell(y,0)$ by $4C(1+|y|^s)$; integrating the third-derivative
bound from zero to $t$ then proves
\eqref{eq:recovery-loss-hessian}, increasing $C_s$ only as a function
of the fixed growth exponent and constant. Convexity makes the
second derivative nonnegative. Thus we may choose
\[
 \mathcal L(R)=\max\{\mu,\lambda B_r(n)
  +C_s M^2(1+C_y(n)^s)(1+(MR)^{s+1})\}.
\]
It is polynomial in $n$ whenever $R$ is polynomial. In particular
$\mathcal L(B_0)$, $G_0$, $R_0$, and $L_+$ are uniformly polynomial
on $\mathcal G_n$. Hence $\underline a^{-1}$ is polynomial,
$j_*=O(\log(en))$, and \eqref{eq:recovery-update-bound} is polynomial
because $\tau_n^{-1}\leq n$.
Every function or gradient evaluation costs $O(np+p)$ in the dense
arithmetic model. Charging all rejected trials, gradient checks,
and the two starting-objective evaluations proves the asserted
$R_{\mathrm{work},n}$ bound. Its constants do not use a noise event.

Couple the guarded process and the process used in the prefix tracking
proof with identical original Gaussian innovations. They have the same
initial state and exactly the same operations until the first possible
recovery. On $E$, at each deletion the first successful gate occurs
within $J_\star=B_{\mathrm N}$ corrections. The gate after the last
allowed correction is checked before the recovery branch, so equality
with the budget also gives acceptance without recovery. Induction
therefore gives identical centers and releases throughout $E$.
Equivalently, a first recovery would require failure of all
$B_{\mathrm N}=J_\star$ post-update gates at that deletion, contradicting
$E$. This proves its probability bound without asserting agreement
after a recovery.

On $E$ the tracking and CG estimates give $W\leq W_{\mathrm{fast},n}$.
On every other finite history, at most $B_{\mathrm N}$ Newton steps per
deletion and at most $p$ exact CG iterations per correction give
$C_B B_{\mathrm N}Tnp^2$ prefix work, including the gradient and residual checks.
At most $T$ recoveries cost $T R_{\mathrm{work},n}$, regardless of how
later histories change. Splitting expectation over $E$ and $E^c$
proves \eqref{eq:guarded-expected-work}. Choose a fixed $d$ larger
than the fixed polynomial degree of its bracket to make that term
lower order; the required sufficiently-large-$n$ threshold may then
increase. This choice changes a proof confidence level, not the
algorithm's tolerance, safeguard or data-dependent behavior.
\end{proof}
The event in Proposition~\ref{prop:guarded-work} must control both actual
gradient-gate acceptance and the CG work of all attempted Newton steps.
For the one-update hidden-state configuration, Appendix~\ref{app:proof-hidden-one-step}
establishes gate acceptance using $\tau_n^{(1)}$ and the common CG residual
rule~\eqref{eq:true-cg-residual}; Proposition~\ref{prop:cg-work} supplies
the associated work bound. A center-distance invariant alone does not
establish the one-update stopping claim. These work statements are
conditional on fixed good data.
They do not establish an expectation over datasets outside
$\mathcal G_n$, a bit-complexity bound, or a floating-point runtime
or termination guarantee.
\clearpage
\section{Experimental Setup}
\label{sec:experimental-details}

\subsection{Data and algorithm settings}
All Newton--CG results use Algorithm~\ref{alg:sequential-newton-cg} with \texttt{is\_perfect=true}. We use summed ridge-logistic loss with penalty $(\lambda/2)\|\beta\|_2^2$, $\lambda=0.1$, $(\epsilon,\delta)=(1,10^{-6})$, and tolerance $\tau_n=1/n$. Starting from a numerical noiseless full-data fit, each deletion uses the preceding published model as its only carried model state. The maximum Newton counts are $J_*=2$ and $3$ for pointwise and pathwise calibration, respectively. CG uses the true-residual threshold $\lambda\tau_n/8$, and the gradient gate $\|\nabla F_{\backslash k}\|_2\leq\lambda\tau_n$ is checked after each update. If the gate fails after $J_*$ updates, Algorithm~\ref{alg:sequential-newton-cg} invokes recovery before publication.

We generate independent covariates $x_i\sim N(0,\Sigma/p)$ and responses $y_i\mid x_i\sim\operatorname{Bernoulli}((1+e^{-x_i^\top\beta_*})^{-1})$, where
\[
 \Sigma=a[I+(\rho-1)vv^\top],\qquad
 a=\frac{p}{p-1+\rho},\qquad
 v=\frac{\mathbf 1}{\sqrt p},\qquad
 \beta_*=\sqrt{p/2}\left(\frac{v}{\sqrt{a\rho}}+\frac{w}{\sqrt a}\right),
\]
with $w=(1,-1,\ldots,1,-1)^\top/\sqrt p$. This gives $\operatorname{tr}(\Sigma)=p$ and unit logit variance; $\rho=1$ gives the isotropic design. Each independent replicate consists of a dataset, a uniform deletion permutation, and publication-noise streams. The noise scales, fixed throughout each path, are
\[
 \sigma_{\mathrm P}=\frac{2\tau_n\sqrt{2\log(2/\delta)}}{\epsilon},
 \qquad
 \sigma_{\mathrm H}=\frac{2\tau_n\sqrt{2T\log(1/\delta)}}{\epsilon}.
\]

\subsection{Computational cost}
We set $n=1200$, $p=T=300$, and $\rho=1000$, with ten independent replicates. Within each calibration, Newton--CG, GD--Armijo, and GD--fixed share data, deletion order, initialization, and standardized publication noise. The GD baselines use their own preceding releases and the same gradient gate and publication rule. For retained design $X_k$, let $B_k=\lambda+\|X_k\|_F^2/4$. GD--fixed uses step $2/(B_k+\lambda)$. GD--Armijo starts each deletion at $1/B_k$, doubles the preceding accepted step up to $1/\lambda$, and halves rejected trials with Armijo constant $10^{-4}$.

We measure cumulative dense linear-algebra FLOPs and online CPU time, including rejected trials and all recovery work; initialization and offline evaluation are excluded. Curves average the ten replicates, with pointwise $95\%$ $t$ intervals. Paired cost ratios are geometric means of within-replicate ratios.

\subsection{Population-risk degradation}
We use $\rho=1$, $n\in\{600,1200,2400,4800\}$, $p=T=n/4$, and twenty independent replicates per $n$. The calibrations share data, deletion order, and initialization but use distinct noise streams. Within each calibration, noisy retraining and Newton--CG share the same publication noise. Numerical noiseless retraining references have gradient residual at most $10^{-10}$ and are used only for offline evaluation. Population logistic risk is evaluated by Gaussian quadrature with refinement and an independent adaptive-integration check at absolute tolerance $10^{-10}$.

For each replicate, we maximize the positive risk difference over all deletions before averaging across replicates, as in Section~\ref{sec:simulations}. Individual $95\%$ percentile intervals use $10{,}000$ paired whole-replicate bootstrap draws within each $n$.

\subsection{Error propagation}
We set $\rho=1$, $n=1200$, and $p=T=300$, with twenty independent replicates. The two calibrations share data, deletion order, initialization, and numerical retraining references, with distinct publication-noise streams. At each deletion, we record distances to the current retraining reference for the entering model, the first complete Newton--CG update, and the final accepted center before publication. Distances are divided by $\tau_n$ and averaged across replicates. Reference fits satisfy the same $10^{-10}$ gradient-residual tolerance and remain outside the online solver. Pointwise $95\%$ intervals use $10{,}000$ whole-replicate bootstrap draws preserving both calibrations and all deletion rounds.

\end{document}